\documentclass{article}

\usepackage{PRIMEarxiv}

\usepackage[utf8]{inputenc} 
\usepackage[T1]{fontenc}    
\usepackage{hyperref}       
\usepackage{url}            
\usepackage{booktabs}       
\usepackage{amsfonts}       
\usepackage{nicefrac}       
\usepackage{microtype}      
\usepackage{lipsum}
\usepackage{graphicx}       
\graphicspath{{media/}}     
\usepackage[font=normalsize,labelfont=bf]{caption}
\usepackage{enumerate}
\usepackage{caption}
\usepackage{algorithm}
\usepackage[noend]{algpseudocode}
\usepackage{hyperref}
\usepackage{diagbox}
\usepackage{xcolor}

\usepackage{blindtext}
\usepackage{setspace}
\hypersetup{colorlinks,allcolors=black}

\usepackage{parskip}

\usepackage{xcolor}

\usepackage{amssymb,amsmath}
\usepackage{stackengine}
\usepackage{enumitem}
\newlist{steps}{enumerate}{1}
\setlist[steps, 1]{wide=0pt, leftmargin=\parindent, label=Step \arabic*:, font=\bfseries}
\usepackage{bm}

\usepackage{algpseudocode}
\usepackage{textcomp}
\usepackage{gensymb}
\usepackage[nottoc]{tocbibind}

\usepackage{subcaption}
\usepackage{natbib}
\setcitestyle{numbers}
\setcitestyle{citesep={,}}
\usepackage{longtable,booktabs,tabularx,tabulary}
\usepackage{multirow}
\usepackage{rotating}
\usepackage{siunitx}
\usepackage[skip=0.33\baselineskip]{caption}
\usepackage{caption} 
\setcitestyle{square}
\newcolumntype{L}[1]{>{\hsize=#1\hsize\raggedright\arraybackslash}X}%
\newcolumntype{R}[1]{>{\hsize=#1\hsize\raggedleft\arraybackslash}X}%
\newcolumntype{C}[2]{>{\hsize=#1\hsize\columncolor{#2}\centering\arraybackslash}X}%

\newcolumntype{Y}{>{\centering\arraybackslash}X}
\algnewcommand\INPUT{\item[\textbf{Input:}]}%
\algnewcommand\OUTPUT{\item[\textbf{Output:}]}%
\newcommand{\AlgBlankLine}{\par\nobreak\vspace*{.5\baselineskip}}
\usepackage[status=draft,index,multiuser]{fixme}
\fxsetface{margin}{\linespread{1}\scriptsize}
\fxusetheme{color}
\usepackage{soul}
\usepackage{algorithm}
\usepackage{enumitem}
\usepackage[tableposition=top]{caption}

\usepackage{authblk}
\usepackage{titlesec}
\titlespacing{\chapter}{20pt}{20pt}{<after-sep>}

\patchcmd{\maketitle}
 {\def\@makefnmark}
 {\def\@makefnmark{}\def\useless@macro}
 {}{}
\title{\textit{FastSVD-ML--ROM}: A reduced-order modeling framework based on machine learning for real-time applications 
}
\author[a,b\thanks{Corresponding author: g.drakoulas@upnet.gr (G. I. Drakoulas)}]{{\large G. I. Drakoulas}}
\author[a]{{\large T. V. Gortsas}}
\author[c]{{\large G. C. Bourantas}}
\author[c]{{\large V. N. Burganos}}
\author[a]{{\large D. Polyzos}}

\affil[a]{Department of Mechanical Engineering and Aeronautics, University of Patras,
Patras, GR-26500, Rion, Greece}
\affil[b]{FEAC Engineering P.C., Patras, GR-26224, Greece }
\affil[c]{Institute of Chemical Engineering Sciences (ICE-HT), Foundation for Research and Technology, Hellas (FORTH), GR-26504 Patras, Greece}
\begin{document}
\maketitle

\vspace{-0.8cm}

\begin{abstract}
Digital twins have emerged as a key technology for optimizing the performance of engineering products and systems. High-fidelity numerical simulations constitute the backbone of engineering design, providing an accurate insight into the performance of complex systems. However, large-scale, dynamic, non-linear models require significant computational resources and are prohibitive for real-time digital twin applications. To this end, reduced order models (ROMs) are employed, to approximate the high-fidelity solutions while accurately capturing the dominant aspects of the physical behavior. The present work proposes a new machine learning (ML) platform for the development of ROMs, to handle large-scale numerical problems dealing with transient nonlinear partial differential equations. Our framework, mentioned as \textit{FastSVD-ML-ROM}, utilizes \textit{(i)} a singular value decomposition (SVD) update methodology, to compute a linear subspace of the multi-fidelity solutions during the simulation process, \textit{(ii)} convolutional autoencoders for nonlinear dimensionality reduction, \textit{(iii)} feed-forward neural networks to map the input parameters to the latent spaces, and \textit{(iv)} long short-term memory networks to predict and forecast the dynamics of parametric solutions. The efficiency of the \textit{FastSVD-ML-ROM} framework is demonstrated for a 2D linear convection-diffusion equation, the problem of fluid around a cylinder, and the 3D blood flow inside an arterial segment. The accuracy of the reconstructed results demonstrates the robustness and assesses the efficiency of the proposed approach.
\end{abstract}

\vspace{0.3cm}


\newenvironment{keyword}{%
 \textbf{\textit{Keywords:}} Reduced order modeling, Machine learning, Parametrized PDEs, Singular value decomposition \raggedright \\ \: \: \: \:  \raggedright }{}
\vspace{0.2cm}
\section{Introduction}
Computational fluid dynamics and structural mechanics are applied amongst others in the field of aerospace \cite{whalen2020hypersonic}, marine \cite{hu2018robust}, bio-engineering \cite{drakoulas2021coupled}, \cite{lampropoulos2022hemodynamics},\cite{bourantas2021immersed}, and energy, \cite{madsen2019multipoint} to simulate with great accuracy, non-linear, time-dependent  physical phenomena. High fidelity models (HFMs) are developed to predict the physical behavior of dynamical systems, as well as, to gain a better insight into the operational performance \cite{gonzalez2018deep}. Hence, the design is examined and significantly optimized before the deployment in the real world \cite{lui2019construction}. Multi-fidelity models are prepared by using various numerical techniques such as the finite element method (FEM), the finite volume method (FVM), the finite difference method (FDM), the meshless collocation methods or the boundary element method (BEM). However, large-scale models with fine mesh grids require significant computational resources \cite{daniel2020model}. Therefore, the use of high-performance computing infrastructure is often essential to simulate large-scale full order models \cite{ahmed2021closures}.

While HFMs are the backbone of the engineering design, the strategies adopted in the fourth industrial revolution for the development and monitoring of large structures, require numerical models able to provide solutions real time \cite{jones2020characterising}. Alongside, the improvement of computer hardware (e.g. multi-core graphics processing units) combined with modern technologies such as cloud applications (e.g., azure) and edge computing, that can handle big data, generates great opportunities for further development of data-driven methods and simulations \cite{rasheed2019digital}. Much of the success has been raised further by artificial intelligence techniques \cite{polyzos2021ensemble} including various deep learning (DL) models applied to discover hidden patterns from noisy data \cite{peng2021multiscale,brunton2020machine,polyzos2021ensemble}. 

Recent works have introduced the concept of digital twins, as a system that integrates physics-based models, with artificial intelligence techniques and data derived from field sensors \cite{botin2022digital, srikonda2020increasing,grieves2019virtually}. Reduced-order models (ROMs) enhance the bridging between the physical and virtual world, providing real-time solutions \cite{rasheed2019digital}. The formulation and application of the ROMs can be found in various multi-query problems such as design optimization \cite{choi2019accelerating}, data assimilation  \cite{pawar2021data, arcucci2019optimal}, and uncertainty quantification \cite{pagani2021enabling, zahr2019efficient}. Reduced-order modeling is a mathematical approach, part of a broader category called surrogate modeling \cite{gonzalez2018deep, benner2015survey}. The objective of the ROMs is to decrease the simulation cost required by the HFM while preserving the dominant solution patterns and predict the spatio-temporal scale of the unknown fields in real time \cite{swischuk2019projection, arzani2021data}. ROMs can be classified into two main categories according to the availability of the HFM operators that describe the dynamics: \textit{i)} intrusive ROMs (IROMs) and \textit{ii)} non-intrusive ROMs (NIROMs). 

Regarding IROMs, the exact form of the underlying partial differential equations (PDEs) is essential to generate simplified models from high fidelity snapshots \cite{kim2021efficient}, in an ordered database called snapshot matrix \cite{gupta2022three}. Proper orthogonal decomposition (POD) approximates the intrinsic dimensions of the HFM solutions by utilizing linear algebra techniques such as principal component analysis, and singular value decomposition (SVD) \cite{pant2021deep,mou2021data}. The galerkin projection methods predict the evolution of the temporal coefficients through the governing equations of the system  \cite{ahmed2020long,ahmed2021multifidelity}. To adress the nonlinear terms of the PDEs, various methodologies have been proposed including the petrov-galerkin (PG) projection \cite{parish2020adjoint} and the least-squares-PG \cite{carlberg2017galerkin}. However, IROMs provide inaccurate results when dealing with complex phenomena such as unsteady, non-linear fluid flows (e.g., turbulence), since the PG-POD approaches do not maintain stability \cite{xie2019non}. The PG method has also proven to be ineffective for capturing a low-dimensional solution subspace in advection-dominated problems \cite{xu2020multi, maulik2021reduced}. Another problem associated with IROMs, is their requirement for access to the solvers \cite{yildiz2021intrusive}, which is often impossible in the industry when commercial softwares are utilised \cite{mucke2021reduced}.

Concerning NIROMs two classes are found: \textit{ii,a)} the POD-based and \textit{ii,b)} the DL-based \cite{kadeethum2022non}. POD-based NIROMs rely mainly on the approximation of a linear trial subspace spanned by a set of basis vectors \cite{fresca2021comprehensive}, and the reconstruction of the HFM solution using mainly machine learning (ML) models \cite{salvador2021non}. Dynamic mode decomposition and sparse identification of nonlinear dynamics have been proposed to manage the spatiotemporal features in a complete data-driven way \cite{ kaheman2020sindy}. A framework based on POD combined with the discrete empirical interpolation method to handle the non-linearities {\cite{saibaba2020randomized}}, and  operator inference {\cite{benner2020operator}} to discover the dynamics, has been introduced in \cite{hartmann2021differentiable}. POD along with gaussian processes have been formulated, to extract the basis vectors of the snapshots, and through regression predict the coefficients of the reduced basis \cite{ortali2022gaussian}. However, existing NIROM-POD based methodologies for dimensionality reduction, rely mostly on a linear subspace \cite{xu2020multi}. In the case of high turbulent channel flows and low Reynolds number, POD requires many basis modes to reconstruct the total energy of the system \cite{hasegawa2020machine}. Therefore, POD-based NIROMs based become ineffective for complex, large-scale problems aiming to provide solutions real-time. 

In recent years, NIROMs DL-based have been the subject of several studies aiming to overcome the limitations of the linear projections methods \cite{gupta2022three}, by recovering non-linear, low-dimensional manifolds \cite{xu2020multi, salvador2021non}. Various methodologies have been proposed, including supervised and unsupervised DL techniques, to identify low-dimensional manifolds and nonlinear dynamic behaviors \cite{nikolopoulos2022non}. A common methodology applied in the frame of NIROMs DL-based is the convolutional autoencoders (CAE) for the non-linear dimensionality reduction, and the long short-term memory (LSMT) networks to predict the temporal evolution \cite{gonzalez2018deep, nakamura2021convolutional}. In \cite{murata2020nonlinear}, mode decomposition is achieved through POD and AE for the nonlinear feature extraction of flow fields. In a recent work \cite{xu2020multi}, the authors utilized CAE for dimensionality reduction, temporal CAE to encode the solution manifold, and dilated temporal convolutions to model the dynamics. In \cite{maulik2021latent}, the performance of the CAE, variational autoencoders and POD to obtain low dimensional embedding has been examined and Gaussian processes regression to implement the mapping between the input parameters and the reduced solutions.

In this work, we propose a generic ROM platform, refereed to as \textit{FastSVD-ML-ROM}, that includes a combination of a SVD update methodology and CAE for the dimensionality reduction. A feed forward neural network (FFNN) and a LSTM network are utilised to predict and forecast the temporal scales of the extracted reduced representations. Our framework presents several novelties compared to existing approaches in the literature: 
\begin{enumerate}[label=(\roman*)]
\item A truncated SVD update \cite{kalantzis2021projection} is employed to calculate the basis vectors of large snapshot matrices during the simulation process, and to identify a low-dimensional embedding. The linear projected data is passed to the CAE, utilized for non-liner dimensionality reduction aiming to compute the latent variables. 
\item A LSTM network is applied to predict and forecast the temporal scales of the latent representations in unseen times for a given parameter vector in real time. To achieve that, a FFNN is employed to map the input parameters sets to the first temporal sequence of the LSTM network during the online phase. 
\item To enhance the generalizabity of the ROM, the input vector of the LSTM network is formulated as a combination of the latent variables and the parameter sets.
\item Finally, the neural architecture search (NAS) is proposed to discover the optimum activation function of FFNN, through a multi-trial exploration strategy \cite{benardos2007optimizing}.
\end{enumerate}

Closest to our approach are the methodologies presented in \cite{fresca2021comprehensive,fatone2022long} and \cite{maulik2021reduced}. Fresca et al. \cite{fresca2021comprehensive} utilized randomized SVD as a prior dimensionality reduction, CAE for non-linear manifold identification and FFNN for the prediction of the dynamic response while Fatone et al. \cite{fatone2022long} proposed a POD-LSTM approach for time extrapolations. Maulik et al. \cite{maulik2021reduced} applied CAE for dimensionality reduction and, LSTM for the prediction of the transient response.  An important difference with \cite{fresca2021comprehensive} is that \textit{FastSVD-ML-ROM} makes predictions and forecasts through the LSTMs, with \cite{fatone2022long} is that CAE are utilized for dimensionality reduction, and, regarding \cite{maulik2021reduced} our framework does not acquire the HFM solution during the online phase to predict the dynamics. To the best of our knowledge, there is no other methodology in the frame of NIROMs, that can handle large-scale HFMs and calculate linear embedding during the simulations process, identify non-linear manifolds, predict and forecast the dynamics for a given parameter during the online phase. 

The structure of this paper is as follows: In section \ref{Machine learning-based reduced order model}, we describe the snapshot matrix construction, the SVD update, and the autoencoders for the linear and non-linear dimensionality reduction, as well. Furthermore, we demonstrate the methodology for the prediction of the dynamics through LSTM networks and the parameter interpolation via the FFNNs. The complete scheme of the \textit{FastSVD-ML-ROM} is described, and the interconnections of the DL models are demonstrated, in section \ref{Complete scheme of the FastSVD-ML-ROM framework}. Then, in section \ref{Numerical Results} we test the developed procedure for three numerical examples: 1) the 2D convection-diffusion case in a square domain, 2) the 2D fluid flow around a cylinder (Navier Stokes equations) and 3) the 3D blood flow in an arterial segment (Navier Stokes equations). Finally, a brief discussions of the developed methodology and results, as well as thoughts about future perspectives are described in section \ref{Conclusions}. 
\section{Machine learning-based reduced order model} \label{Machine learning-based reduced order model}
In this section, the setting of the transient numerical problems as well as the basic components of the proposed \textit{FastSVD-ML-ROM} platform are addressed. More precisely, to efficiently compress the input data, a hyper-reduction technique based on both linear algebra (SVD) and DL techniques by utilizing CAE, is developed. Furthermore, a combination of FFNN and LSTM models is utilized, to predict the spatio-temporal evolution of the reduced solution. The linear algebra methodology and the DL models are described in the following sections, to gain a better insight on the architecture of the ROM methodology. 

\subsection{Problem setting}
The general form of a time-dependent, parameterised, non-linear PDE can be expressed as follows:
\begin{equation} \label{eq:1}
    \partial_t{\bm{u}}(t,\bm{x};\bm{\mu})=\mathcal{F}(t,\bm{x};\bm{u};\bm{\mu}), \  \ \bm{u}(0,\bm{x};\bm{\mu})=\bm{u}_0(\bm{x};\bm{\mu}),  \  \ t\in (0,T],  \  \ \bm{x}\in \Omega
\end{equation}
where, $\Omega \subset \mathbb{R}^d$ is the spatial domain of interest, $\bm{u}$: $\mathbb{R}^{+} \times \Omega \times \mathbb{R}^{\xi}$ $\rightarrow$ $\mathbb{R}^{g}$ is the unknown vector function, $\mathcal{F}$ is a non-linear differential operator, $\bm{u}_0$: $\mathbb{R}^d \times \mathbb{R}^{\xi}$ $\rightarrow$ $\mathbb{R}^{g}$ is the initial condition, $\bm{x}$ $\in$ $\Omega$ denotes a spatial point, $(0, T]$ is the temporal domain, $\xi$ is the dimension of the parameter vector and $d$ is the number of spatial dimensions. The well-posedness of the problem (Eq.\ref{eq:1}), requires a certain number of initial and boundary conditions, that must be satisfied so that the problem to have unique solutions on $\partial{\Omega}$. The parameter vector \bm{$\mu$} is sampled from a parameter space $\mathcal{P}$ $\subset$ ${\mathbb{R}}^{\xi}$, that constitutes the space of interest for the HFM solution, and it may consist for example of boundary or initial conditions (e.g. the inlet velocity) and material properties \cite{kadeethum2022non}. 

Numerical methods provide an efficient set of tools to approximate the solution of Eq.\ref{eq:1}. The FEM \cite{pettas2015origin} is a well established computational method for solving differential equations in various applications in engineering. The FVM \cite{eymard2000finite} and FDM \cite{liszka1980finite} have been  extensively applied in the field of fluid mechanics, heat and mass transfer \cite{moukalled2016finite}, while the BEM \cite{rodopoulos2019aca,sellountos2020single} is an ideal method to solve linear problems, in infinite and semi-infinite domains \cite{GORTSAS2017108,kalovelonis2020cathodic}. Meshless collocation methods constitute a significant tool, alleviating the mesh burden in computational mechanics \cite{bourantas2019explicit}. In recent years, physics-informed neural networks \cite{raissi2019physics} have been proven an efficient way to solve the forward and inverse problems using DL methods.

In the present work, the FVM, FEM and meshless collocation methods are utilised to compute the HFM solution of Eq. \ref{eq:1} for the numerical examples, examined in section \ref{Numerical Results}. The spatially discretized form of Eq. \ref{eq:1} can be represented in the following form,  

\begin{equation} \label{PDE-dis}
    \partial_t\bm{u}_h(t;\bm{\mu})=\mathcal{F}_h(t,\bm{u}_h;\bm{\mu}),  \   \  \bm{u}_h(0; \bm{\mu})=\bm{u_0}(\bm{\mu}),   \   \  t\in (0,T] 
\end{equation}
where $h>$0 denotes a discretization parameter, such as the element size. The HFM parameterised discrete solution is denoted as $\bm{u}_h(t;\bm{\mu})$: $\mathbb{R}^+ \times \mathbb{R}^{\xi}$ $\rightarrow$ $\mathbb{R}^{N_h}$, and $N_h$ the dimension of the discretized function $\bm{u}$, which can be extremely large for complex, multi-physics simulations in realistic geometries. The parameters $\bm{\mu}$ are uniformly sampled from the space $\mathcal{P}$ and are split in two sets $\bm{\mu}^{tr}$=$\{\bm{\mu}_1^{tr},\ldots,\bm{\mu}_m^{tr}\}$ and $\bm{\mu}^{te}$=$\{\bm{\mu}_1^{te},\ldots,\bm{\mu}_n^{te}\}$ corresponding to the $m$ training and $n$ testing parameters, respectively, where
$\mu^{tr}\cup\mu^{te} \subset{\mathcal{P}}$. The spatial discrete solution of the high-fidelity model satisfying Eq. \ref{PDE-dis}, is computed for the parameter vector that contains both the training parameters ($\bm{\mu}^{tr}$), that construct the snapshot matrix, and the testing parameters ($\bm{\mu}^{te}$) that examine the accuracy of the $\textit{FastSVD-ML-ROM}$. 
The snapshot matrix $\bm{S}_h$ is defined for the parameter set $\bm{{\mu}}^{tr}$ and the discrete time set $\{{t_1,\ldots,t_{N_t}}\}$ with $t_i$ $\in$ $(0,T]$, as follows,
\begin{equation} \label{eq:snap}
    \bm{S}_h = [\; \bm{u}_h(t_1;\bm{\mu}^{tr}_1) \;| \ldots| \; \bm{u}_h(t_{N_t},\bm{\mu}^{tr}_1) \; |\ldots| \; \bm{u}_h(t_{1},\bm{\mu}^{tr}_m) \;|\ldots| \; \bm{u}_h(t_{N_t},\bm{\mu}^{tr}_m) \;]
\end{equation}
where $\bm{u}_h(t_i,\bm{\mu}_j)$ is a snapshot for the parameter $\bm{\mu}_j^{tr}$ at time $t_i$.
\subsection{Linear dimensionality reduction} \label{section:SVD}
To efficiently compress the above-defined snapshot matrix (Eq. \ref{eq:snap}), the truncated SVD update algorithm \cite{kalantzis2021projection} is exploited. By utilizing an SVD update methodology, we manage to compress big snapshot matrices during the simulation process, in a reduced time compared to the classical SVD.

\begin{figure}[t] 
\centering\includegraphics[width=1\linewidth]{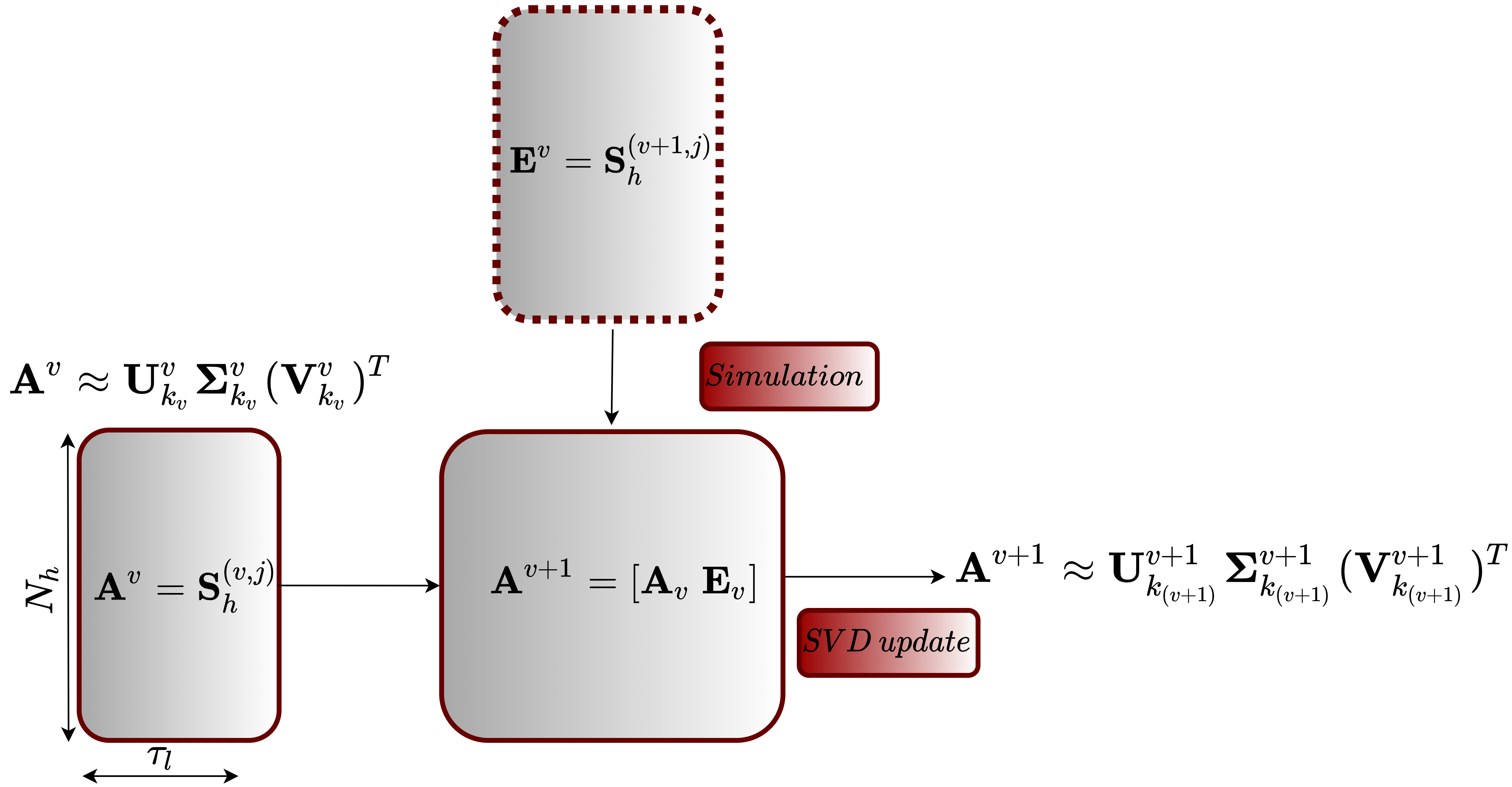}
\caption{Illustration of the truncated SVD update during the simulation phase. The HFM solution $\bm{E}^{v}$ proceeds to the SVD algorithm to calculate the updated basis vectors $\bm{U}_{k_{v+1}}^{v+1}$, $\bm{V}_{k_{v+1}}^{v+1}$, and the singular vector $\bm{\Sigma}_{k_{v+1}}^{v+1}$ of the composed matrix $\bm{A}^{v+1}$. The methodology is repeated for $\forall$ $\bm{\mu}^{tr}_j$ with $j=1,\ldots,m \in \mathcal{P}$ and mini-batches ${\tau}_v$ with $v=1,\ldots,l$.\label{SVD_up}}
\end{figure}

The first step is the partitioning of the computed HFM solutions $\bm{u}_h(t_i;\bm{\mu}^{tr}_j)$ in small clusters. Each cluster corresponds to the solutions obtained in a certain time interval. The solutions in each interval are calculated from the HFM incrementally during the compression procedure. Then, the SVD is computed for the part of the snapshot matrix that corresponds to the solution for the first parameter $\bm{\mu}^{tr}_1$ and for the first time interval. The computed SVD is continuously updated considering the solution obtained for consequent time intervals and the other parameter values $\bm{\mu}_j^{tr}$, with $j$=$1,\ldots,m$. Following this approach, a reduced basis is calculated during the simulation process, that approximately captures the range of the snapshot matrix (Fig. \ref{SVD_up}). 

We consider that the discrete time set $\{t_1,...,t_{N_t}\}$ with $ t_{i} \in (0,T]$ is partitioned in $l=\frac{N_t}{s}$ subsets as follows:
\begin{equation} \label{data set}
   \{ t_1,...,t_{N_t} \} =  {\tau}_{1} \cup  {\tau}_{2} \cup \ldots {\tau}_{l}
\end{equation}
where ${\tau}_{v}= \{t_{(v-1)\cdot s+1},\ldots,t_{v\cdot s} \}$ with $v=1,\ldots,l$ and s is a predefined parameter which controls the efficiency of the proposed algorithm. 

Based on this partition, the sub-matrix of $\bm{S}_h$ corresponding to the parameter $\bm{\mu}_{j}^{tr}$ and the subset ${\tau}_{v}$ is written as:
\begin{equation} \label{eq:snapshot matrix}
    \bm{S}_h^{(v,j)} = [\;\bm{u}_h(t_{(v-1)s+1};\bm{\mu}^{tr}_j) \;| \ldots| \;\bm{u}_h(t_{v\cdot s};\bm{\mu}_j^{tr})\;]
\end{equation}

For the sake of brevity, the adopted algorithm is described for $\bm{S}_h^{(v,1)}$, corresponding to the first parameter, since the same procedure will be repeated for the results obtained for all the other parameters. \\ Initially, the truncated SVD of the matrix $\bm{A}^{1}=\bm{S}_h^{(1,1)} \in \mathbb{R}^{N_hxs}$ is computed as:
\begin{equation}
    \bm{A}^1\approx \bm{U}^1_{k_1} \bm{\Sigma}^1_{k_1} ({\bm{V}^1_{k_1}})^T
\end{equation}
where $\bm{U}^1_{k_1} \in \mathbb{R}^{N_h \times k_1}$ and $\bm{V}^1_{k_1} \in \mathbb{R}^{s \times k_1}$ are matrices which contain the left and right $\mathit{k_1}$ singular vectors, respectively, and $\bm{\Sigma}^1_{k_1} \in \mathbb{R}^{k_1 \times k_1}$ is a diagonal matrix which contains the positive singular values $\bm{\sigma}$. The rank $\mathit{k_1}$ is computed such that
$\bm{\sigma}_{k_{1}}\leq\varepsilon\bm{\sigma}_1$, where $\varepsilon$ is a predefined accuracy. 

During the simulation process we are interested in computing the updated SVD of the evolving matrix $\bm{A}^{v+1}= [\bm{S}_h^{(v,1)} \ \ \ \bm{S}_h^{(v+1,1)}]=[\bm{A}^v \ \ \ \bm{E}^{v}]$, with $v\leq{l-1}$. The truncated SVD of the matrix $\bm{A}^{v}$ is assumed to be known and $\bm{E}^{v} \in \mathbb{R}^{N_h \times s}$ is the simulation result obtained for the sub-interval $\boldsymbol{\tau}_{v+1}$.

We first consider the projector $\bm{\Pi}^v={\bm{U}^{v}_{k_v}}({\bm{U}^{v}_{k_v}})^T$ with $\bm{\Pi}^v  \in$ $\mathbb{R}^{N_h \times N_h}$ and compute the QR decomposition \cite{goodall199313} of the matrix $(\bm{I}-\bm{\Pi}^v)\bm{E}^v$ as follows,
\begin{equation}
     (\bm{I}-\bm{\Pi}^v)\bm{E}^v = \bm{Q}^v\bm{R}^v 
\end{equation}
where $\bm{Q}^v\in \mathbb{R}^{N_h \times s}$ is orthonormal, and $\bm{R}^v \in \mathbb{R}^{s \times s}$ is an upper triangular matrix.  
Using the above expression, $\bm{E}^v$ can be written in the following form, 
\begin{equation}
    \bm{E}^v = \bm{Q}^{v}\bm{R}^{v}+{\bm{U}^{v}_{k_v}}({\bm{U}^{v}_{k_v}})^T\bm{E}^{v}=
    \left[
    \begin{array}{cc} 
    {\bm{U}^{v}_{k_v}} & \bm{Q}^{v} 
    \end{array}
    \right]
    \left[
    \begin{array}{cc} 
	\bm{0} & ({\bm{U}^{v}_{k_v}})^T\bm{E}^v \\
	\bm{0} & \bm{R}^{v}
	\end{array}
	\right]
    \left[
    \begin{array}{cc} 
	\bm{0} \\
	\bm{I}
	\end{array}
	\right]
\end{equation}
This yields the following factorization,
\begin{equation} \label{eq:2}
    \bm{A}^{v+1}\approx
    \left[
    \begin{array}{cc} 
    {\bm{U}^{v}_{k_v}} \bm{\Sigma}^v_{k_v} {( {\bm{V}^{v}_{k_v}})}^T & 
    \bm{Q}^v\bm{R}^v+{\bm{U}^{v}_{k_v}}({\bm{U}^{v}_{k_v}})^T\bm{E}^v 
    \end{array}
    \right]=
    \left[
    \begin{array}{cc} 
    {\bm{U}^{v}_{k_v}} & \bm{Q}^v
    \end{array}
    \right]    
    \left[
    \begin{array}{cc} 
    \bm{\Sigma}^v_{k_v} & ({\bm{U}^{v}_{k_v}})^T\bm{E}^v \\ 
    \bm{0} & \bm{R}^v
    \end{array}
    \right] 
    \left[
    \begin{array}{cc} 
    {( {\bm{V}^{v}_{k_v}})}^T & \bm{0} \\
    \bm{0} & \bm{I}
    \end{array}
    \right] 
\end{equation}
where $\bm{I}$ is the $k_{v} \times k_{v}$ identity matrix. The key is then to compute the SVD of the ${(k_{v}+s)\times(k_{v}+s)}$ matrix,
\begin{equation}
   \left[
    \begin{array}{cc} 
     \bm{\Sigma}^v_{k_v} & ({\bm{U}^{v}_{k_v}})^T\bm{E}^v \\ 
    \bm{0} & \bm{R}^v
    \end{array}
    \right] = \bm{F}^v\bm{\varTheta}^v(\bm{G}^v)^T
\end{equation}    
so that Eq. \ref{eq:2} obtains the form: 
\begin{equation}
    \bm{A}^{v+1}\approx
   \left[
    \begin{array}{cc} 
    {\bm{U}^{v}_{k_v}} & \bm{Q}^v 
    \end{array}
    \right]
    \bm{F}^v\bm{\varTheta}^v{(\bm{G}^v)}^T
    \left[
    \begin{array}{cc} 
     (\bm{V}^{v}_{k_v})^T & \bm{0} \\
    \bm{0} & \bm{I}
    \end{array}
    \right]  
\end{equation}    
Setting,
\begin{equation}
   {\bm{U}}^{v+1}=
   \left[
    \begin{array}{cc} 
    {\bm{U}^{v}_{k_v}} & \bm{Q}^v 
    \end{array}
    \right]
    \bm{F}^v, \;
    {\bm{V}}^{v+1}=
    \left[
    \begin{array}{cc} 
     {( {\bm{V}^{v}_{k_v}})}^T & \bm{0} \\
    \bm{0} & \bm{I}
    \end{array} 
    \right]^T{\bm{G}^v}  
\end{equation}  
we obtain the SVD of the matrix $\bm{A}^{v+1}$ as, 
\begin{equation}
    \bm{A}^{v+1} \approx {\bm{U}}^{v+1} \, \bm{\varTheta}^v \, ({\bm{V}}^{v+1})^{^T}
\end{equation}
The rank of the matrix $\bm{A}^{v+1}$ can be finally truncated to $k_{v+1}$ so that  $\bm{A}^{v+1} \approx  \bm{U}^{v+1}_{k_{v+1}} \, \bm{\varTheta}^v \, (\bm{V}^{v+1}_{k_{v+1}})^T$. The truncated SVD update procedure is applied to the solutions obtained for all the time subsets $\tau_{v}$ with $v=1,\ldots,l-1$ and sequentially for each parameter $\bm{\mu}_{j}^{tr}$ of the HFM, with $j=1,\ldots,m$. Following this approach, we obtain a matrix $\bm{U}$ with orthogonal columns and with rank $n = k_{m \cdot l}$ such that $range(\bm{S}_h) \approx range(\bm{U})$. It is worth mentioning that the compression is efficient when $n<<N_t$ and $n<<N_h$, while the projection error $\varepsilon$ is chosen small enough to obtain the required accuracy in the reconstructed solution. Using the calculated basis an approximated solution is obtained as:
\begin{equation}\label{eq14}
    \Tilde{\bm{u}}_h(t_i;\bm{\mu}^{tr}_j) \approx {\bm{U}} \ \bm{U}^T \bm{u}_h(t_i;\bm{\mu}_j^{tr})
\end{equation} 
Given the snapshot matrix $\bm{S}_h$ (Eq. \ref{eq:snap}) containing the solutions for all the parameters, we can calculate a low dimensional projection as follows,
\begin{equation} \label{eq15}
    {\bm{S}}_n =  {\bm{U}}^T{\bm{S}}_h
\end{equation}
where the projected matrix ${\bm{S}}_n$ is the input given to the CAEs and n the column dimensions of the basis vector ${\bm{U}}$ derived during the SVD update. Therefore, the projected solution is derived through the formula, ${\bm{u}_n(t_i,\bm{\mu}_j^{tr})}={\bm{U}}^T\bm{u}_h(t_i;\bm{\mu}^{tr}_j)$.
\vspace{-5mm}
\subsection{Non-linear dimensionality reduction}
\begin{figure}[t]
\centering\includegraphics[width=1\linewidth]{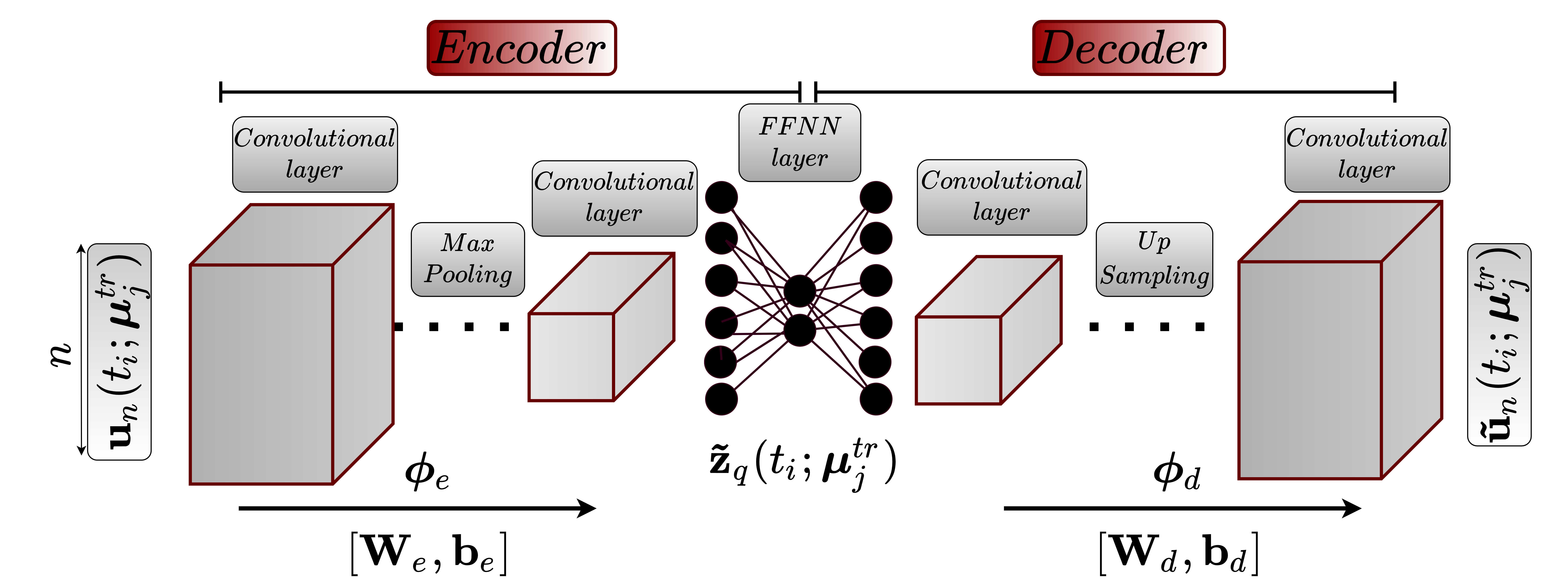}
\caption{CAE architecture including the encoder and the decoder parts. Starting from the linear low approximation of the HFM ${\bm{u}_n}(t_i; \bm{\mu}^{tr}_j)$ to the latent space $\mathbf{\tilde{z}}_q(t_i;\bm{\mu}^{tr}_j)$ and back to the reconstructed solution $\bm{\tilde{u}}_n(t_i; \bm{\mu}^{tr}_j)$. \label{AE}}
\end{figure}

After decreasing the dimensions of ${\bm{S}}_h$ through the truncated SVD (Eq. \ref{eq15}), we use the projected solutions $\bm{u}_n(t_i;\bm{\mu}^{te}_j)$ as input in the AE network to identify a non-linear, low-dimensional manifold for the reduced solution. In general, the AE is an unsupervised type of neural network (NN), that approximately learns an identity mapping of the unlabeled input data. It is commonly used for matrix factorization, dimensionality reduction and principal component analysis \cite{aggarwal2018neural} as well as, for image compression and feature detection \cite{mao2016image}. AEs map the input tensor to a latent space through the encoder denoted as $e$, and then the decoder denoted as $d$ transforms the latent representation (Fig. \ref{AE}). 

Let $\bm{\varphi}_e$ be the encoder function that attempts to discover a latent representation  $\mathbf{\tilde{{z}}}_q$ $\in$ $\mathbb{R}^q$, with $q<<n$,
\begin{equation}
     \mathbf{\Tilde{{z}}}_q(t_i;  \bm{\mu}^{tr}_j)=\bm{\varphi}_e(\bm{u}_n(t_i;  \bm{\mu}^{tr}_j);\bm{W}_e;\bm{b}_e), \:\:\:  with \:\:\: i=1,\ldots,N_t \:\:\: and \:\:\: j=1,\ldots,m
\end{equation}
while $\bm{W}_e$, $\bm{b}_e$ are groups of weight matrices and bias vectors of the encoder, respectively. 

The decoder reconstructs the approximate input data $\bm{\Tilde{u}_{n}}(t_i;\bm{\mu}_j^{tr})$ through the following transformation,
\begin{equation} \label{Eq:17}
    \bm{\Tilde{u}}_n(t_i;\bm{\mu}^{tr}_j)=\bm{\varphi}_d( \mathbf{\Tilde{{z}}}_q(t_i;\bm{\mu}^{tr}_j);\bm{W}_d;\bm{b}_d)=\bm{\varphi}_d(\bm{\varphi}_e(\bm{u}_n(t_i;{\bm{\mu}^{tr}_j});\bm{W}_e;\bm{b}_e);\bm{W}_d;\bm{b}_d)
\end{equation}
where the transition functions are defined as $\bm{\varphi}_e: {\mathbb{R}^n} \rightarrow {\mathbb{R}^q}$ and $\bm{\varphi}_d: {\mathbb{R}^q} \rightarrow {\mathbb{R}^n}$, while
$\bm{W}_d$, $\bm{b}_d$ are the weight matrices and bias vectors of the decoder. In the frame of \textit{FastSVD-ML-ROM}, to efficiently handle high dimensional data and capture the spatial scales, convolutional AEs (CAEs) are utilised \cite{yildirim2018efficient}. The main idea of the CAEs is the utilization of filters to reduce the rank of $\bm{S}_n$. 

We consider the function of the encoder $\bm{\varphi}_e$ that takes as input the reduced solution $\bm{u_n}(t_i;\bm{\mu}_j^{tr})$. In the first step, the data is reshaped in a square matrix, applying zero padding, if necessary. Then, convolutional layers apply a kernel through a moving window to encode the data. The filtered output is then normalized, and downsampled through max-pooling operations. The procedure is repeated multiple times to compress the data, and FFNN are used to generate the latent space representations, $\mathbf{\Tilde{{z}}}_q(t_i;\bm{\mu}^{tr}_j)$.

In the inverse direction, the function of the decoder $\bm{\varphi}_d$ transforms the reduced solutions following a similar architecture, by converting the latent spaces in a vector through FFNN, followed by convolutional layers and upsampling layers to generate the reconstructed output $\mathbf{\tilde{u}}_n(t_i;\bm{\mu}_j^{tr})$. For the sake of generality we denote as $\bm{W}_e, \bm{b}_e$ and $\bm{W}_d, \bm{b}_d$ the weight matrices and bias vectors of the CAEs. A detailed explanation of the CAE architecture is presented in \cite{nikolopoulos2022non}.

We train the CAE by minimizing the loss function $\mathcal{L}^{c}$, among the input data $\bm{u}_n(t_i;\bm{\mu}^{tr}_j)$, and the reconstructed data $\bm{\Tilde{u}}_n(t_i,\bm{\mu}^{tr}_j)$, with $i=0,\ldots,N_t$ and $j=1,\ldots,m$.
\begin{equation} \label{Loss_CAE}
    \mathcal{L}^{c} = \frac{1}{m \cdot N_t } \sum_{j=1}^{m}\sum_{i=1}^{N_t}||\bm{u}_n(t_i;\bm{\mu}_j^{tr})-\bm{\Tilde{u}}_n(t_i,\bm{\mu}_j^{tr})||^2
\end{equation}
During the training, the number of times that the data is passed thought the network is controlled by the epochs number $e_c$. The loss function function $\mathcal{L}^{c}$ is calculated and the parameters are updated over each iteration using multiple training samples, determined by the batch size aiming to speed-up the completion of every epoch. 
\subsection{Prediction of dynamics and forecasting}
To model the evolution of the latent variables extracted from the HFM snapshots, time series prediction models are used. Amongst others, recurrent NNs have been proposed in the DL community to describe the evolution of nonlinear dynamics \cite{song2019air}. LSTMs are a modified version of recurrent NNs, developed to overcome the stability issues of traditional time series models and in particular the phenomenon of vanishing gradients\cite{mohan2018deep}. LSTMs have been successfully applied to predict linear or non-linear sequential data, as well as to forecast, accounting for past and current observations \cite{venugopalan2015sequence}. The major advantage of the LSTM network, is the ability to efficiently learn the evolution of the fields using gating mechanisms to control the flow of information, as part of their architecture \cite{zhang2021dive}. 

We denote the LSTM network, as $l$ that takes as input the vector $\bm{x}_l^i(\bm{\mu}_j^{tr})$, consisting of the latent spaces $\mathbf{\tilde{z}}_q(t_i,\bm{\mu}_j)$ with $i=1,\ldots,N_t$ and $j=1,\ldots,m$  extracted through the trained encoder $\bm{\varphi}_e$. To explicitly account for the parameter information (e.g. velocity, material properties, etc), we concatenate the latent spaces with the parameter values, and thus $\bm{x}_l^i(\bm{\mu}^{tr}_j)=[\mathbf{\tilde{z}}_q(t_i;\bm{\mu}_j^{tr}), \: \bm{\mu}_j^{tr}]$. 

In particular, multiple samples are generated through a predefined sliding window $w$, arranged in a temporal sequence to enter the LSTM network. Each input set is passed incrementally for every parameter $\bm{\mu}^{tr}_j$ as follows,
\begin{equation}
    \mathcal{X}_l(\bm{\mu}_j^{tr}) = \{ \; \bm{X}_l^1(\bm{\mu}_j^{tr}),\ldots,\bm{X}_l^g(\bm{\mu}_j^{tr}) \; \}
 \end{equation}
where $g=(N_t-w)$, and each matrix $\bm{X}_l^i(\bm{\mu}_j^{tr})$ contains w column vectors, given as, 
\begin{equation}
    \bm{X}_l^i(\bm{\mu}_j^{tr})=[ \; \bm{x}_l^i(\bm{\mu}_j^{tr}) \;| \ldots| \; \bm{x}_l^{(i+w-1)}(\bm{\mu}_j^{tr})\; ]
\end{equation}
The LSTM network is trained by the data $\bm{X}_l^i(\bm{\mu}_j^{tr})$ to predict the next latent space vector $\bm{y}_l^i(\bm{\mu}_j^{tr})=\mathbf{\tilde{z}_q}(t_{i+w};  \: \bm{\mu}_j^{tr})$.
Given that LSTM model is deployed for all matrices  $\bm{X}_l^i(\bm{\mu}_j^{tr})$, the outputs of the training sequence for each parameter $\bm{\mu}_j^{tr}$ form the following set $\mathcal{Y}_l$,
\begin{equation}
    \mathcal{Y}_l(\bm{\mu}_j^{tr})=\{\bm{y}_l^1(\bm{\mu}_j^{tr}),\ldots,\bm{y}_l^g(\bm{\mu}_j^{tr}) \}
\end{equation}

The LSTM network mainly consists of three gates acting as filters to the input data at time $t_i$ and parameter $\bm{\mu}_j$: the forget gate $\bm{f}_i(\bm{\mu}^{tr}_j)$ that decides which information is less important and can be ignored, the input gate $\bm{i}_i(\bm{\mu}^{tr}_j)$ used to update the cell state and the output gate $\bm{o}_i(\bm{\mu}^{tr}_j)$ that determines the information that will be passed to the next state. Besides, the LSTM cell, includes the update cell state $\bm{C}_i(\bm{\mu}^{tr}_j)$, composed by the input gate $\bm{i}_i(\bm{\mu}^{tr}_j)$, the forget gate $\bm{f}_i(\bm{\mu}^{tr}_j)$ and the candidate memory state $\bm{\tilde{C}}_i(\bm{\mu}^{tr}_j)$ that generates data. The hidden state $\bm{h}_i(\bm{\mu}^{tr}_j)$ produces the output given the filtered data from the output gate $\bm{o}_i(\bm{\mu}^{tr}_j)$ \cite{yan2015understanding}. Then, the hidden state is passed to a FFNN layer aiming to predict the final output data $\bm{\tilde{y}}_l^1(\bm{\mu}_j^{tr})$.
\begin{figure}[t]
\centering\includegraphics[width=1\linewidth]{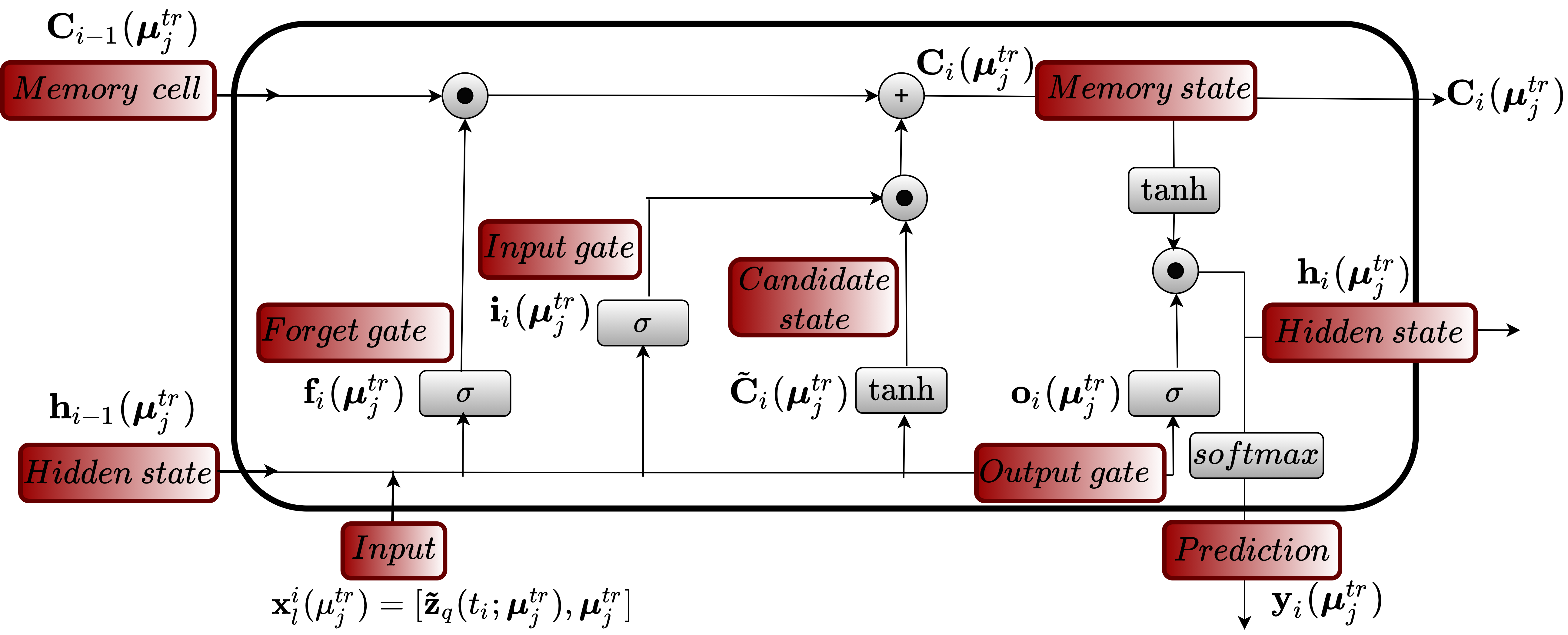}
\caption{The architecture of the LSTM network, as part of the \textit{FastSVD-ML-ROM} platform. The input vector $\bm{x}_l^i$ includes the values of the latent spaces and the parameter values in a concatenated vector,  $\bm{x}_l^i(\mu_j^{tr})=[\mathbf{\tilde{z}}_q(t_i; \bm{\mu}_j^{tr});  \: \bm{\mu}_j^{tr}]$.}
\end{figure}
The basic formulas of an LSTM cell are described in detail below.

The forget gate, $\bm{f}_t(\bm{\mu}^{tr}_j)$ constitutes the first component of the LSTM cell. The input vector of the forget gate $\bm{f}_i(\bm{\mu}^{tr}_j)$, is formulated through a concatenation of the previous hidden state $\bm{h}_{i-1}(\bm{\mu}^{tr}_j)$ and the input vector $\bm{x}_l^i(\bm{\mu}_j^{tr})$.
\begin{equation}
\bm{f}_i(\bm{\mu}^{tr}_j)=\mathit{\sigma}(\bm{W}_{fo}[\bm{h}_{i-1}(\bm{\mu}_j^{tr}), \: \bm{x}_l^i(\bm{\mu}_j^{tr})] + \bm{b}_{fo})
\end{equation}

Then, through the first term of the memory cell state $\bm{C}_i(\bm{\mu}^{tr}_j)$ (Eq. \ref{candidate}), the network removes the less important information.
\begin{equation} \label{candidate}
    \bm{C}_i(\bm{\mu}^{tr}_j) = \bm{f}_i(\bm{\mu}^{tr}_j)\odot\bm{C}_{i-1}(\bm{\mu}^{tr}_j)+\bm{i}_i(\bm{\mu}^{tr}_j)\odot\Tilde{\bm{C}_i}(\bm{\mu}^{tr}_j)
\end{equation}
Apart from the usage of forgetting mechanisms,  the LSTM architecture includes filters utilised to retain and generate data.  
In particular, the LSTM network, through the second term of the updating cell state $\bm{C}_i(\bm{\mu}^{tr}_j)$ (Eq. \ref{candidate}), creates an update to the current state. 

The candidate memory cell state $\Tilde{\bm{C}_i}(\bm{\mu}^{tr}_j)$ adds new state values in Eq. \ref{candidate}, using the following function:
\begin{equation}
    \mathbf{\Tilde{C}}_i(\bm{\mu}^{tr}_j) = tanh(\bm{W}_{ca}[\bm{h}_{i-1}(\bm{\mu}^{tr}_j), \: \bm{x}_l(\mu_j^{tr})] + \bm{b}_{ca})
\end{equation}
and the input of the LSTM network is controlled by the input gate $i_i$, expressed as,
\begin{equation}
    \bm{i}_i(\bm{\mu}^{tr}_j)=\mathit{\sigma}(\bm{W}_{in}[\bm{h}_{i-1}, \: \bm{x}_l^i(\mu_j^{tr})] + \bm{b}_{in})
\end{equation}
The output data is achieved in three phases. First, the output gate $\bm{o}_t$ computes the generated data as,
\begin{equation}
    \bm{o}_i(\bm{\mu}^{tr}_j) = \mathit{\sigma} (\bm{W}_{out} [\bm{h}_{i-1}, \: \bm{x}_l^i(\bm{\mu}^{tr}_j)(\mu_j^{tr})] + \bm{b}_{out}]
\end{equation}
Then, the hidden state $\bm{h}_i$ acting as filter, determines the data passed to the next iteration,
\begin{equation}
    \bm{h}_i(\bm{\mu}^{tr}_j) = {\bm{o}_i}\odot\tanh({\bm{C}_i})
\end{equation}
Finally, a FFNN employed with a softmax activation function, is applied to create the final output $\mathbf{\tilde{y}}_i$ of the LSTM layer such as,
\begin{equation}
    \mathbf{\tilde{y}}_i(\bm{\mu}_j^{tr})=softmax(\bm{W}_y\bm{h}_i(\bm{\mu}^{tr}_j) + \bm{b}_y)
\end{equation}

where, $\bm{W}_{in}$, $\bm{W}_{fo}$, $\bm{W}_{ca}$, $\bm{W}_{out}$ and $\bm{W}_{y}$ are the weight matrices of the input gate ($in$),the  forget gate ($fo$), the candidate state ($ca$), the output gate ($out$) and the FFNN ($y$) that formulate the weight matrix of the LSTM network denoted as $\bm{W}_l=[\bm{W}_{in},\bm{W}_{fo},\bm{W}_{ca},\bm{W}_{out},\bm{W}_{y}]$. The terms $\bm{b}_{in}, \bm{b}_{fo}, \bm{b}_{ca}$, $\bm{b}_{out}$ and $\bm{b}_y$ construct a vector including the bias of each layer such as, $\bm{b}_l$=[$\bm{b}_{in},\bm{b}_{fo},\bm{b}_{ca},\bm{b}_{out},\bm{b}_{y}$]. Besides, $\sigma$ is the logistic sigmoid function, $tanh$ is the hyperbolic tangent function, $softmax$ is the normalized exponential function \cite{nwankpa2018activation}, while $\odot$ is the element-wise product.

We denote as $\bm{\varphi}_l$ the general function of each LSTM cell that attempts to predict the next step given incrementally the input data $\bm{X}_l^i(\bm{\mu}_j^{tr})$,
\begin{equation} \label{Eq:23}
    \mathbf{\tilde{y}}_l^i(\bm{\mu}_j^{tr}) = \bm{\varphi}_l(\bm{X}_l^i(\bm{\mu}_j^{tr});\bm{W}_l;\bm{b}_l)
\end{equation}
where $i$ = 1, \ldots ,$g$ and $j$ = 1, \ldots ,$m$. The LSTM network is trained $e_l$ epochs and the final loss function of the LSTM network $\mathcal{L}^l$, aims to be minimized is given by,
\begin{equation}
    \mathcal{L}^l = \frac{1}{g \cdot m } \sum_{j=1}^{m}\sum_{i=1}^{g}||\bm{y}_l^i(\bm{\mu}_j^{tr})-\mathbf{\tilde{y}}_l^i(\bm{\mu}_j^{tr})||^2
\end{equation}
 
The major advantage of the LSTM network is the ability to forecast (extrapolate) to un-seen times, out of the simulation time-range, and in particular for time \(t > T\), where $T$ is the duration of the ROM training.
\subsection{Parameter regression} \label{FFNN}
FFNNs, also known as multi-layer perceptrons, approximate the functional relation among a set of input and output values. The FFNN is composed by multiple layers with a variable number of nodes, defining a mathematical sequence of operations that consists of affine transformations, followed by element wise multiplications of a (non)linear activation function.

\begin{figure}[H]
\centering\includegraphics[width=1\linewidth]{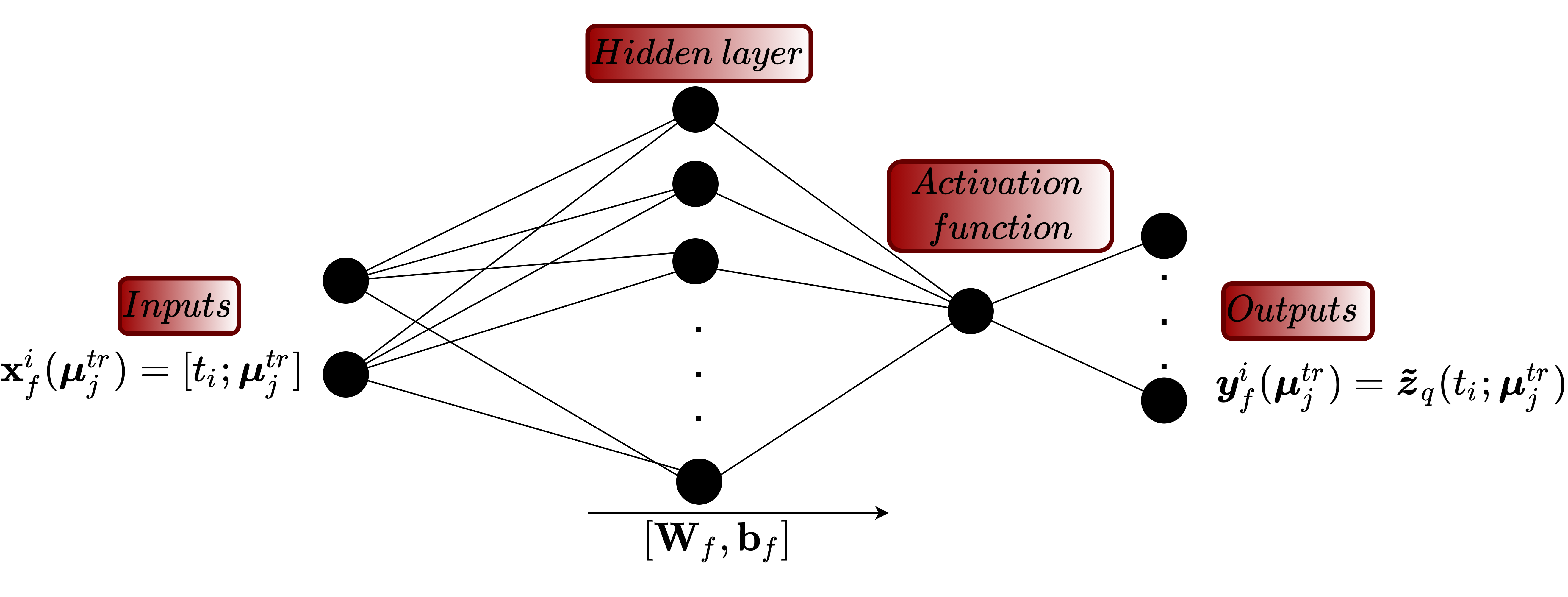}
\caption{The architecture of the FFNN, as part of the \textit{FastSVD-ML-ROM} framework. \label{FFNN_Ar}}
\end{figure}

In the frame of \textit{FastSVD-ML-ROM}, FFNNs are used to identify the latent space vectors  $\bm{\Tilde{{z}}}_{\bm{q}}(t_i;\bm{\mu}_j^{tr})$ for $i=1,\ldots,w$ and $j=1,\ldots,m$ that form the first time series matrix  \textbf{$\mathbf{X}_l^1(\bm{\mu}^{tr}_j)$, $j=1,\ldots,m$}, provided as input to the LSTM during the online phase to predict and forecast the dynamics (Fig. \ref{FFNN_Ar}).

To formulate the input for the FFNNs, we concatenate the time instances $t_i$, with $i=1,\ldots,w$ with the parameters $\bm{\mu}_j^{tr}$, and thus $\bm{x}_f^i(\bm{\mu}_j^{tr})=[t_i, \: \bm{\mu}_j^{tr}]$. 
In particular the training set of the FFNN is the following,
\begin{equation}
    \bm{x}_f(\mu_j^{tr})=\{\bm{x}_f^1(\mu_j^{tr}),\ldots,\bm{x}_f^w(\mu_j^{tr})\}
\end{equation}

The output of the FFNN for a given parameter $\mu_j^{tr}$ is the following set
\begin{equation}
    \bm{y}_f(\mu_j^{tr})=\{\bm{y}_f^i(\bm{\mu}^{tr}_j),\ldots,\bm{y}_f^w(\bm{\mu}^{tr}_j) \}
\end{equation}
where, $\bm{y}_f^i(\bm{\mu}^{tr}_j)=\bm{\Tilde{{z}}}_{\bm{q}}(t_i;\bm{\mu}_j^{tr})$.

In particular, we aim to obtain the nonlinear mapping such that: 
\begin{equation} \label{Eq:27}
    \mathbf{\tilde{y}}^i_f(\mu_j^{tr})=\bm{\varphi}_f(\bm{x}_f^i(\bm{\mu}_j^{tr});\bm{W}_f,\bm{b}_f)
\end{equation}
Where $\bm{b}_f$ are the weight and bias terms of the FFNN and $\bm{\varphi}_f:\mathbb{R}^{n_{\bm{\mu}}+1}$ $\rightarrow$ $\mathbb{R}^q$ is a (non)linear activation function.
The final loss function ($\mathcal{L}^f$) of the FFNN is as follows, 
\begin{equation}
    \mathcal{L}^f = \frac{1}{m \cdot w } \sum_{j=1}^{m}\sum_{i=1}^{w}\left|\left|\bm{y}_f^i(\bm{\mu}^{tr}_j)-\mathbf{\tilde{y}}^i_f(\bm{\mu}_j^{tr})\right|\right|^2
\end{equation} 
The FFNN is trained for $e_f$ epochs aiming to minimize the loss function ($\mathcal{L}^f$).

In the frame of the \textit{FastSVD-ML-ROM} framework, the non-linear activation function of the FFNN is approximated by implementing the NAS technology \cite{elsken2019neural}. A model search space is utilised to identify the best function that maps the input data to the output in an optimum way. In particular, a multi-trial strategy is used to explore the various models that are trained independently. The performance estimation of each activation function is examined through the minimum error obtained during the minimization of the loss function $\mathcal{L}^f$. 
\section{Complete scheme of the proposed \textit{FastSVD-ML-ROM} platform} \label{Complete scheme of the FastSVD-ML-ROM framework}
\begin{figure}[t] 
\centering
\includegraphics[width=1\linewidth]{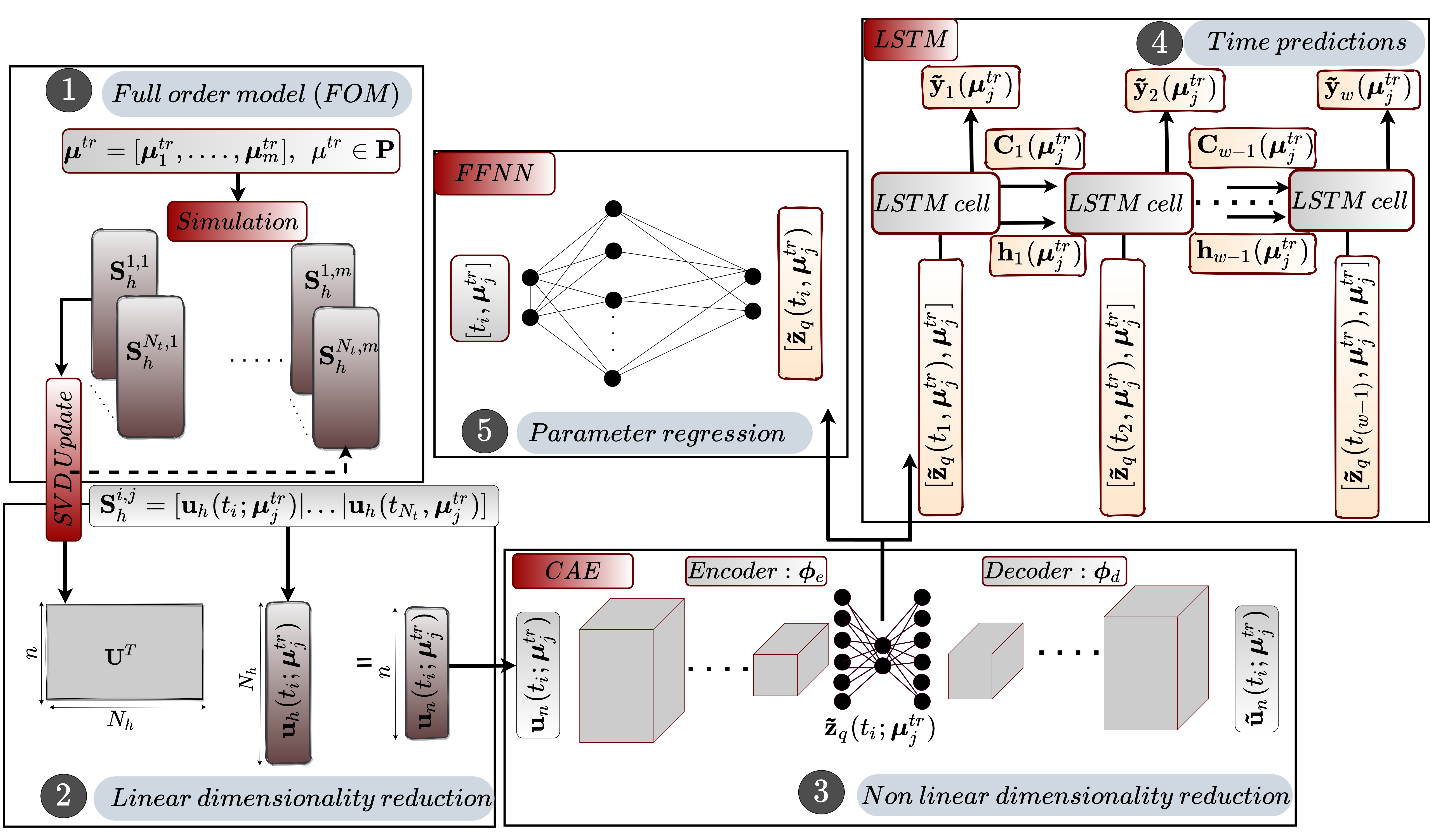}
\caption{Offline phase (training) of the \textit{FastSVD-ML-ROM}. \label{offline fig}}
\end{figure}

In this section, we summarize the operation of the \textit{FastSVD-ML-ROM} framework during the offline (training) and online (testing) phase. In the offline phase, the truncated SVD calculates the basis matrix $\bm{U}$ by giving incrementally partitions of the snapshot matrix $\bm{S}_h$. Then, the linearly projected data $\bm{S}_n$ (Eq. \ref{eq15}) is passed to the DL models, which are independently trained. In particular, the CAE is used for the non-linear dimensionallity reduction, the FFNN for the parameter regression and the LSTM for time predictions. The operations of the offline phase are implemented in Algorithm \ref{offline} and are schematically explained in Fig. \ref{offline fig}.

\begin{figure}[H]
\centering
\includegraphics[width=0.97 \linewidth]{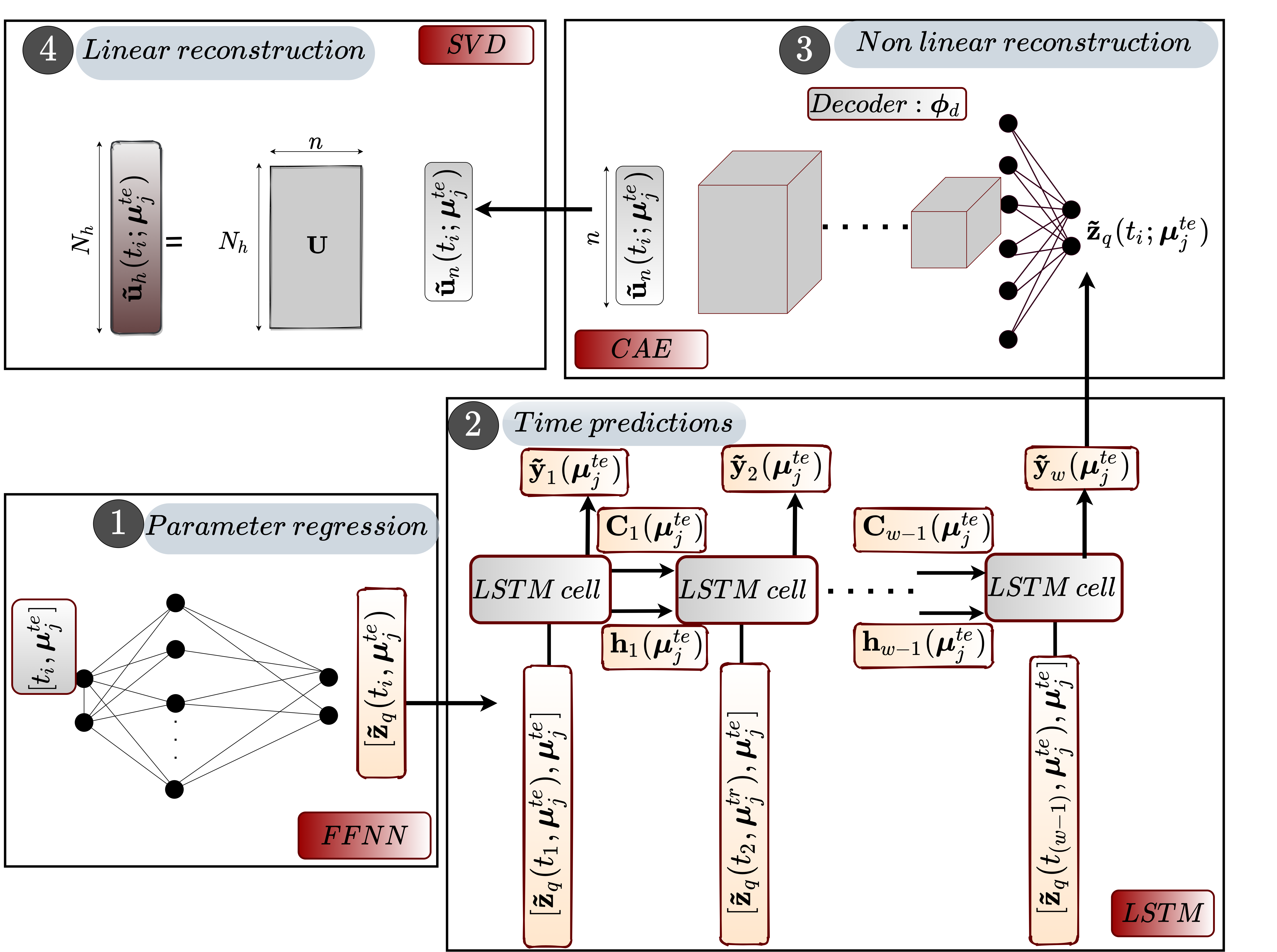}
\caption{Online phase (testing) of the \textit{FastSVD-ML-ROM}. \label{online fig}}
\end{figure}

In the online phase the input parameter vector $\bm{\mu}^{te}$ is passed to the trained FFNN, which obtains the first time sequence of the latent variables (Eq. \ref{Eq:27}) and passes the data to the LSTM network that predicts the evolution of the solution field (Eq. \ref{Eq:23}).  Finally, the decoding part of the CAE (Eq. \ref{Eq:17}) is employed along with the basis matrix $\bm{U}$ (Eq. \ref{eq14}) to calculate the approximated solution given the latent representations in a limited amount of time. The operations of the online phase are implemented in Algorithm \ref{online} and illustrate in Fig. \ref{online fig}.

\vspace{-7.7mm}
\begin{algorithm}[t]
\caption{Offline phase (training)}\label{offline}
  \begin{algorithmic}[1]
    \INPUT number of training parameters $m$, number of testing parameters $n$, parameter space $\mathcal{P}$, time range $t$, timesteps $N_t$, number of epochs $e_c$ for CAE, $e_l$ for LSTM and $e_f$ for FFNN, truncation error $\varepsilon$, number of subsets $l$ for the truncated SVD update.
    \OUTPUT $\bm{\mu}^{tr}$, $\bm{\mu}^{te}$, $\bm{U}$, and the optimal parameters of the LSTM $[\bm{W}_l, \bm{b}_l],$, the FFNN $[\bm{W}_f, \bm{b}_f]$, and the CAE $[\bm{W}_e,\bm{W}_d, \bm{b}_e,\bm{b}_d]$.
    \AlgBlankLine
    \State Sample $\bm{\mu}^{tr}_j$, with $j=1,\ldots,m$ and $\bm{\mu}^{te}_{j}$ with $j=1,\ldots,n$ parameters from $\mathcal{P}$, thought the LHS method.   
    \For {$j=1$ to $m$}
        \For {$v=1$ to $l$}
            \If {($j=1$ and $v=1$)}
                \State Compute the HFM $S_h^{(1,1)}$ and apply SVD 
            \Else
                \State Compute the HFM $\bm{S}_h^{(v,j)} = [\;\bm{u}_h(t_{(v-1)s+1};\bm{\mu}^{tr}_j) \;| \ldots| \;\bm{u}_h(t_{v\cdot s};\bm{\mu}_j^{tr})\;]$ and apply SVD update 
            \EndIf
    \EndFor 
    \EndFor
    \State Calculate the projected snapshot matrix $\mathbf{{S}}_n=\bm{U}^T\bm{S}_h$ using the basis matrix $\bm{U}$
   \For {$k=1$ to $e_c$}
   \For {$j=1$ to $m$}
   \For {$i=1$ to $N_t$}
        \State $\mathbf{\Tilde{{z}}}_{\bm{q}}(t_i;\bm{\mu}^{tr}_j)=\bm{\varphi}_e(\bm{u}_n(t_i;\bm{\mu}^{tr}_j);\bm{W}_e;\bm{b}_e)$ 
        \State $\bm{\Tilde{u}}_n(t_i;\bm{\mu}^{tr}_j)=\bm{\varphi}_d( \mathbf{\Tilde{{z}}}_q(t_i;\bm{\mu}^{tr}_j);\bm{W}_d;\bm{b}_d)$
        \State Calculate loss function $\mathcal{L}^c(\bm{u}_n(t_i;\bm{\mu}^{tr}_j), \: \bm{\Tilde{u}}_n(t_i;\bm{\mu}^{tr}_j))$ and update the parameters through $ADAM$
   \EndFor
   \EndFor
   \EndFor
   \State Form the LSTM set of inputs, $\mathcal{X}_l(\bm{\mu}_j^{tr}) = \{ \; \bm{X}_l^1(\bm{\mu}_j^{tr}),\ldots,\bm{X}^g(\bm{\mu}_j^{tr}) \; \}$  
   \State Form the LSTM set of outputs, $\mathcal{Y}_l(\bm{\mu}_j^{tr})=\{\bm{y}_l^1(\bm{\mu}_j^{tr}),\ldots,\bm{y}_l^g(\bm{\mu}_j^{tr}) \}$
   \For {$k=1$ to $e_l$}
   \For {$j=1$ to $m$}
   \For {$i=1$ to $g$}
        \State $\mathbf{\tilde{y}}_l^i(\bm{\mu}_j^{tr})$ = $\bm{\varphi}_l(\bm{X}_l^i(\bm{\mu}_j^{tr}); \bm{W}_l; \bm{b}_l)$
        \State Calculate loss $\mathcal{L}^l(\bm{y}_l^i(\bm{\mu}_j^{tr}), \: \mathbf{\tilde{y}}_l^i(\bm{\mu}_j^{tr}))$ and update the parameters through $ADAM$
    \EndFor
    \EndFor
    \EndFor
    \State Form the FFNN  set of inputs, $\bm{x}_f(\bm{\mu}_j^{tr})=\{\bm{x}_f^1(\bm{\mu}_j^{tr}),\ldots,\bm{x}_f^w(\bm{\mu}_j^{tr})\}$
    \State Form the FFNN set of outputs,   $\bm{y}_f\bm{(\mu}_j^{tr})=\{\bm{y}_f^i(\bm{\mu}^{tr}_j),\ldots,\bm{y}_f^w(\bm{\mu}^{tr}_j)$
    \For {$k=1$ to $e_f$}
    \For {$j=1$ to $m$}
    \For {$i=1$ to $w$}
             \State   $\mathbf{\tilde{y}}^i_f(\bm{\mu}_j^{tr})=\bm{\varphi}_f(\bm{x}_f^i; \bm{W}_f; \bm{b}_f )$
            \State Calculate the loss $\mathcal{L}^f(\bm{y}_f^i(\bm{\mu}_j^{tr}), \: \mathbf{\tilde{y}}_f^i(\bm{\mu}_j^{tr}))$ and update the parameters through $ADAM$
    \EndFor
    \EndFor
    \EndFor
  \end{algorithmic}
\end{algorithm}
\vspace{5pt}

\begin{algorithm}[H]
\caption{Online phase (testing)}\label{online}
  \begin{algorithmic}[1]
    \INPUT  $\bm{U}$, $\bm{\mu}^{te}$, and the optimal parameters $\bm{W}_l, \bm{b}_l, \bm{W}_f, \bm{b}_f, \bm{W}_e,\bm{W}_d, \bm{b}_e,\bm{b}_d$.
    \OUTPUT  ROM approximation $\mathbf{\tilde{u}}_h(t_i;\bm{\mu}^{te}_j)$.
    \AlgBlankLine
    \State Form  $\bm{x}_f^i(\bm{\mu}_j^{te})=[t_i,  \bm{\mu}_j^{te}]$ with $i=1,\ldots,w$
    \State Calculate $\mathbf{\tilde{y}}^i_f(\bm{\mu}_j^{te})=\bm{\varphi}_f(\bm{x}_f^i(\mu_j^{te});\bm{W}_f;\bm{b}_f)$ through FFNN
    \State Form $\bm{x}_l^i(\bm{\mu}^{te}_j)=[\mathbf{\tilde{y}}^i_f(\bm{\mu}_j^{te}), \bm{\mu}_j^{te}]$ and $\bm{X}_l^i(\bm{\mu}_j^{te})=[ \; \bm{x}_l^i(\bm{\mu}_j^{te}) \;| \ldots| \; \bm{x}_l^{(i+w-1)}(\bm{\mu}_j^{te})\; ]$
    \State Calculate $\mathbf{\tilde{y}}_l^i(\bm{\mu}_j^{te}) = \bm{\varphi}_l(\bm{X}_l^i(\bm{\mu}_j^{te});\bm{W}_l; \bm{b}_l)$ through LSTM 
    \State Reconstruct $\bm{\Tilde{u}}_n(t_i;\bm{\mu}^{te}_j)=\bm{\varphi}_d(\mathbf{\tilde{y}}_l^i(\bm{\mu}_j^{te}) ;\bm{W}_d;\bm{b}_d)$ through the decoder of the CAE
    \State  Reconstruct the $\bm{\Tilde{u}}_h(t_i;\bm{\mu}^{te}_j)=\bm{U}\bm{\Tilde{u}}_n(t_i;\bm{\mu}^{te}_j)$
   \end{algorithmic}
\end{algorithm}

Regarding the minimization of the loss function achieved through the backpropagation algorithm, we utilize the adaptive moment estimation (ADAM) optimizer \cite{kingma2014adam}, a version of stochastic gradient descent \cite{ketkar2017stochastic} that employs first and second moment estimates to derive adaptive learning rates for various parameters. All parameters of the DL-models in the frame of \textit{FastSVD-ML-ROM} are tuned through a hyperparameter optimization analysis \cite{yang2020hyperparameter}. 

\setlength{\parskip}{3pt plus2pt}
\section{Numerical results} \label{Numerical Results}
Three benchmark problems are employed to assess the performance of the proposed \textit{FastSVD-ML-ROM} framework, namely, \textit{i)} a linear convection diffusion equation in a square domain (section \ref{CD}), \textit{ii)} the flow around a cylinder (section \ref{Cylinder}) and \textit{iii)} the blood flow in an arterial segment (section \ref{Arterial}). The DL models are implemented in tensorflow 2.3 \cite{abadi2016tensorflow} and the SVD technique executed by the numpy library \cite{oliphant2006guide} in python 3.8.3. NAS is applied through the neural network intelligence library provided by Microsoft \cite{nni2021}, to discover the optimum function of the FFNN that maps the parameters to the latent space representations. The performance of the developed NAS methodology for each test case is presented in Appendix \ref{NAS}. The numerical simulations and the training of the DL-models are executed on an Intel Core i7 @ 2.20GHz, NVDIA GeForce GTX 1050 Ti GPU personal computer. 
\begin{spacing}{0.3}
\end{spacing}
The accuracy of the developed surrogate model is examined for the testing parameters $\bm{\mu}_j \in \{\bm{\mu}_1^{te},\ldots,\bm{\mu}_n^{te}\}$, at $t_i$, with i=1,\ldots,$N_t$, through one field and three scalar error indicators: \newline
the absolute error ${\varepsilon_{abs}}$ $\in \mathbb{R}$ defined as,
\begin{equation}
                     {\varepsilon_{abs}} =  ||\mathbf{{u}}_h(t_i;\bm{\mu}_j^{te})-\mathbf{\tilde{u}}_h(t_i;\bm{\mu}_j^{te})||
            \end{equation}
the relative error ${\varepsilon_{rel}}$ $\in \mathbb{R}$ defined as,
\begin{equation}
             {\varepsilon_{rel}} = \frac{\sum_{i=1}^{N_h}||{\mathbf{u}_h(t_i;\bm{\mu}_j^{te})-\mathbf{\tilde{u}}_h(t_i;\bm{\mu}_j^{te}})||}{\sum_{i=1}^{N_h}||\mathbf{u}_h(t_i;\bm{\mu}_j^{te})||}
\end{equation} 
the normalised root mean squared error ${\varepsilon_{nrms}}$ $\in \mathbb{R}$ is defined as,
\begin{equation}
        {\varepsilon_{nrms}} = 
        \frac{ \sqrt{\frac{\sum_{i=1}^{N_h}{(\bm{u}_h(t_i;\bm{\mu}_j^{te})-\mathbf{\tilde{u}}_h(t_i;\bm{\mu}_j^{te}))}^2}{N_h}} }{\bm{u}_h(t_i;\bm{\mu}_j^{te})^{max}-\bm{u}_h(t_i;\bm{\mu}_j^{te})^{min}}
    \end{equation}
and the l2-norm error ${\varepsilon_{l2}}$ $\in \mathbb{R}$ to examine the truncated SVD update defined as,
\begin{equation}
        {\varepsilon_{l2}} = 
        \frac{\sqrt{\sum_{i=1}^{N_h}({\mathbf{u}_n(t_i;\bm{\mu}_j^{te})-\mathbf{\tilde{u}}_n(t_i;\bm{\mu}_j^{te}}))^2}}{\sqrt{\sum_{i=1}^{N_h}\mathbf{u}_n(t_i;\bm{\mu}_j^{te})^2}}, 
    \end{equation}

To improve the training of the proposed ROM, data standardization is applied on the training input vectors of CAEs $\bm{u}_n(t_i,\bm{\mu}_j^{tr})$ with $\bm{\mu}_j \in \{\bm{\mu}_1^{tr},\ldots,\bm{\mu}_m^{tr}\}$, while the extracted latent space representations $\mathbf{\tilde{z}}_q(t_i,\bm{\mu}_j^{tr})$ are normalized between 0 and 1, using the following formulas,
\begin{equation}
    Standardization:\bm{u}_n(t_i;\bm{\mu}_j^{tr}) \longmapsto \frac{\bm{u}_n(t_i;\bm{\mu}_j^{tr})-\bm{u}_n(t;\bm{\mu}^{tr})_{mean}}{\bm{u}_n(t;\bm{\mu}^{tr})_{std}}
\end{equation}
\begin{equation}
    Normalization:\mathbf{\tilde{z}}_q(t_i;\bm{\mu}_j^{tr}) \longmapsto \frac{\mathbf{\tilde{z}}_q(t_i;\bm{\mu}_j^{tr})-\mathbf{\tilde{z}}_q(t;\bm{\mu}^{tr})_{min}}{\mathbf{\tilde{z}}_q(t;\bm{\mu}^{tr})_{max}-\mathbf{\tilde{z}}_q(t;\bm{\mu}^{tr})_{min}}
\end{equation}
where the terms of the standardization are derived through the following formulas,
\begin{equation}
    \bm{u}_n(t,\bm{\mu}^{tr})_{mean} = \frac{1}{N_t \cdot m}{\sum_{i=1}^{N_t}\sum_{j=1}^{m}\bm{u}_n(t_i,\bm{\mu}_j^{tr})} \:\:\: and \:\:\:  \bm{u}_n(t,\bm{\mu}^{tr})_{std} =  \sqrt{\frac{\sum_{i=1}^{N_t}\sum_{j=1}^{m}\left(\bm{u}_n(t_i,\bm{\mu}_j^{tr})-\bm{u}_n(t,\bm{\mu}^{tr}\right)_{mean})}{N_t \cdot m}} 
\end{equation}
and the terms of the normalization,
\begin{equation}
    \mathbf{\tilde{z}}_q(t;\bm{\mu}^{tr})_{max} = \max_{i \: \in \: (1,\ldots,N_t)} \max_{j \: \in \:  (1,\ldots,m)} \mathbf{\tilde{z}}_q(t_i,\bm{\mu}_j^{tr}) \:\:\: and \:\:\:  \mathbf{\tilde{z}}_q(t;\bm{\mu}^{tr})_{min} = \min_{i \: \in \: (1,\ldots,N_t)} \min_{\: j \: \in (1,\ldots,m)} \mathbf{\tilde{z}}_q(t_i,\bm{\mu}_j^{tr}) 
\end{equation}
\vspace{-8mm}
\subsection{Benchmark problem 1: 2D unsteady convection-diffusion in a rectangular space}  \label{CD}
As a first example we consider the following convection-diffusion equation \cite{morton2019numerical}:
\begin{equation} 
   \label{eq:CD}
   \frac{\partial c (\bm{x},t)}{\partial t} = \varepsilon \nabla^2 c (\bm{x},t) + \bm{\upsilon} \cdot \nabla c (\bm{x},t) 
\end{equation}
defined in a bounded domain $\Omega := \bm{x} = (x, \: y) \in [0 \: ,1]^2\: m$ and subjected to the boundary conditions \cite{gortsas2022local},
\begin{align} 
    \begin{array}{cc}
    \begin{cases}
         & c(x,y,t)|_{x=0} = \frac{1}{1+4t} exp\left({{-\frac{(-\upsilon_xt-0.5)^2 + (y-\upsilon_yt-0.5)^2}{(1+4t)}}} \right) \\
         & c(x,y,t)|_{x=1} = \frac{1}{1+4t} exp\left({{-\frac{(1-\upsilon_xt-0.5)^2 + (y-\upsilon_yt-0.5)^2}{(1+4t)}}} \right) \\
         & c(x,y,t)|_{y=0} = 0 \\
         & \frac{\mathit{d} c}{\mathit{d} y}(x,y,t)|_{y=1}=0
        \end{cases}
    \end{array}
\end{align}
and to the initial condition \cite{gortsas2022local},
\begin{equation} 
    c (x,y,0) = exp\left({-(x-0.5)^2-(y-0.5)^2}\right)
\end{equation}
with, $c(\bm{x},t)$ being the concentration field at $\bm{x} = (x,y)$ at time $t$, $\varepsilon$ being the diffusivity coefficient, $\nabla$ and $\nabla^2$ being the gradient and Laplace operator, respectively, $\bm{\upsilon}$=$(\upsilon_x,\upsilon_y)$ being the constant velocity vector of the convection term in $x-$ and $y-$direction, as well. 

The governing equation is solved for 24 different velocity vectors, while $m=20$ are considered for training and $n=4$ for testing. Thus, the resulting vector of the DL-model $\bm{\mu}^{tr} = [\bm{\upsilon}_1,...,\bm{\upsilon}_{20}]$ with
$\bm{\upsilon}_i \in [100 , \: 200]^2$\: m/s. The parameters $\bm{\mu}^{tr}$ are uniformly sampled using the latin hypercube sampling method \cite{stein1987large}. We numerically solve the governing equations using the meshless point collocation method \cite{BOURANTAS20131117}. The differential operators are approximated using the discretization corrected particle strength exchange method \cite{BOURANTAS2016285}. The spatial domain is discretized using a set of uniformly distributed nodes with a resolution of $h = 0.005$ to endure a grid while ensuring a grid independent numerical solution, resulting to 160,801 nodes. An explicit time integration scheme is used to model the transient term with a time step $ dt = 10^{-4} \: s$ (the time step ensures the stability of the explicit solver). We set the final time for the simulation to $T = 0.0075$ s, resulting to $N_t$= 75 time solutions for each parameter. \newline 
To test the efficiency of the meshless coloration method, the numerical results are compared with the analytical solution given as \cite{al2018novel},
\begin{equation}
 c(x,y,t) = \frac{1}{1+4t}exp \left({-\frac{(x-\upsilon_xt-0.5)^2+(y-\upsilon_yt-0.5)^2}{(1+4t)}}\right)
\end{equation}

\begin{figure}[t]
\centering\includegraphics[width=0.9\linewidth]{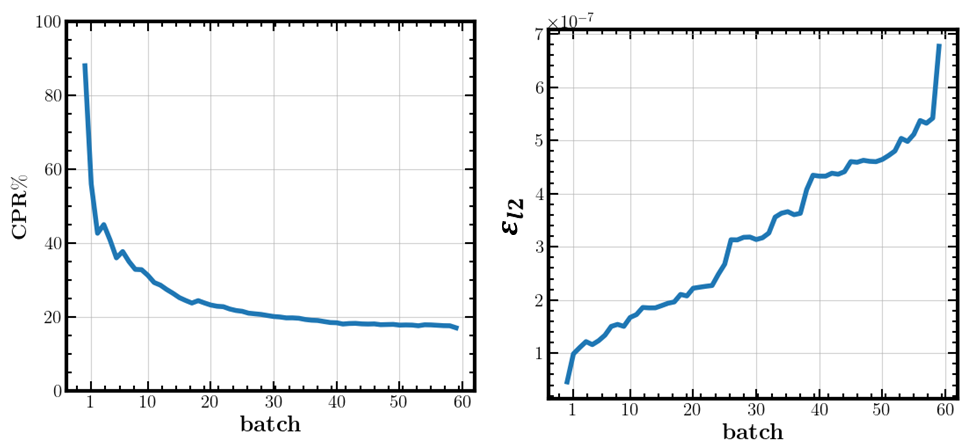}
\caption{The implementation of the truncated SVD update for the CD test case. The evolution of the compression ration (CPR) (left) and the $\epsilon_{l2}$ (right) are presented versus the batches using a truncation error $\epsilon$ = $10^{-7}$. \label{SVD CD}}
\end{figure}

To construct the low-dimensional basis $\bm{U}$ appearing in Eq. \ref{eq15}, we partition the HFM time solutions for each $\bm{\mu}_j^{tr}$ with $j=1$,$\ldots$,$m$ in $l=3$ subsets, aiming to calculate the SVD of the evolving matrices in a reasonable time. Thus, every subset contains $s$ = 25 time solutions. For the updated SVD, a predefined accuracy $\varepsilon=10^{-7}$ is utilised, resulting in a basis of size $n = 256$. To gain a better insight on the performance of the linear projection, we define the compression ratio (CPR) as the  fraction of the reduced basis dimension to the rank of the original snapshot matrix. The results of the CPR and the accuracy of the compression are presented for the CD case in Fig. \ref{SVD CD}. It is shown that the CPR decays rapidly, exhibiting approximately an exponential behavior. During the SVD update of the $1^{st}-20^{th}$ batch file, the CPR presents a significant decay rate, with the maximum compression being close to 20\%. During the SVD update of the $30^{th}-60^{th}$ batch, there is not a significant improvement in the CPR. This is attributed to the higher values of the velocities $\bm{\upsilon}$, which correspond to the latter batches. Alongside, the evolution of the $\varepsilon_{l2}$ error is presented as a function of time, indicating that the error increases for the parameters corresponding to higher velocity values.

\begin{figure}[t]
 \centering
\includegraphics[width=1\linewidth]{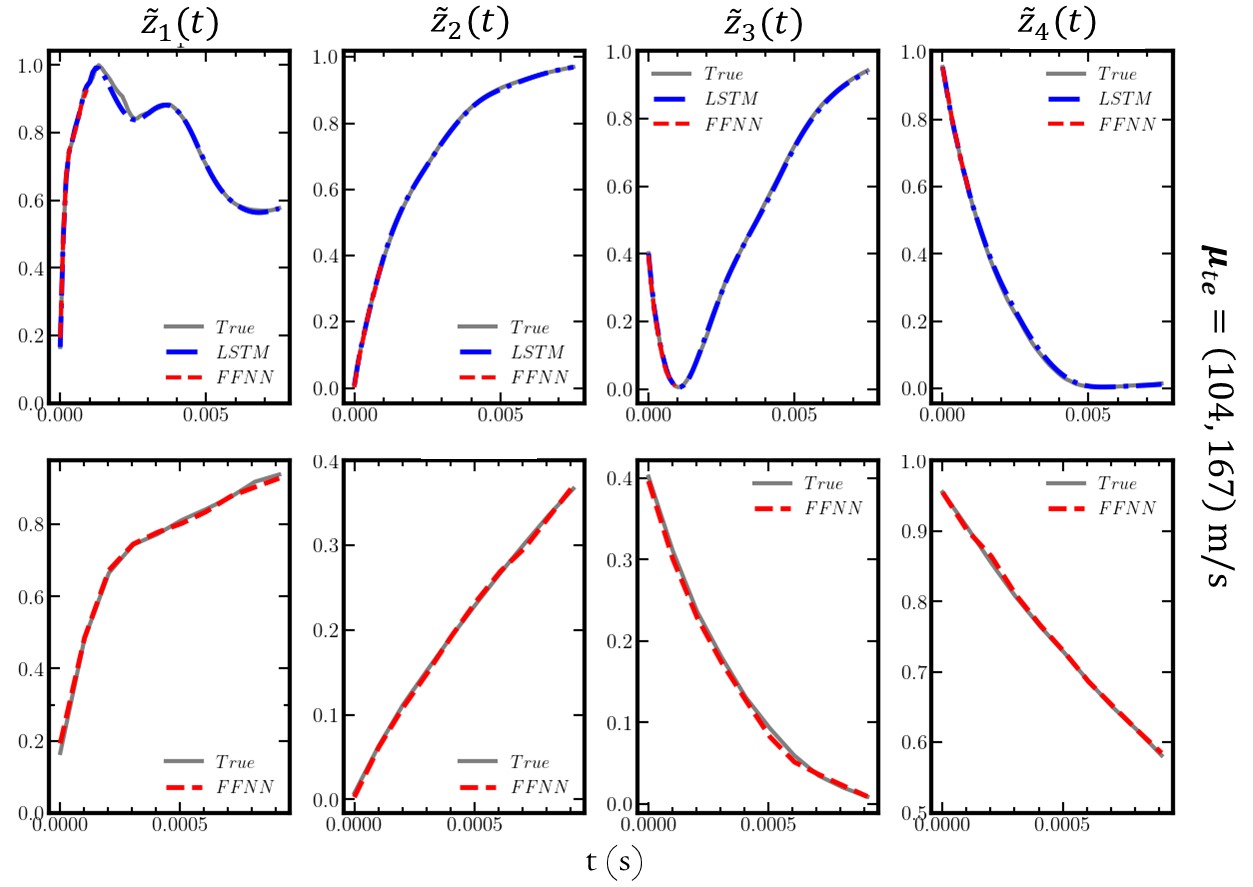}
\caption{True latent space representations extracted by the trained decoder, the LSTM and the FFNN predictions (top) at $t$=[0, 0.0075] s, and the FFNN predictions (bottom) at $t$=[0, 0.001] s for the testing parameter $\bm{\mu}^{te}$=(104, 167) m/s. \label{fig: CD-LSTM}}
\end{figure}
\setlength{\parskip}{0pt}
Regarding the CAE, we set a latent space dimension, $q=4$, so that the input is compressed to $1.56\%$ of the initial linear basis size ($n = 256$). To predict the dynamics, we utilize an LSTM network with a time window $w=10$. For each DL-model, a detailed description of the NN-configurations including the training and validation losses, the NN parameters, and the computational time are presented in Table \ref{Table: CD}. To demonstrate the ability of the NIROM framework to provide the full scale solutions in real-time, we define the speed-up number of each DL-model as the fraction of the testing time and the computational time of the HFM simulation. As concerns the implementation of the NAS (Appendix \ref{NAS}), the Leaky ReLU activation function is utilised, presenting the minimum validation and training loss.

To gain a better insight in the performance of the \textit{FastSVD-ML-ROM}, each DL-model is tested separately. Firstly, we examine the ability of the LSTM and FFNN models to capture the dynamics of the CD equation (Fig. \ref{fig: CD-LSTM}). To achieve that, the latent space representation extracted by the trained decoder, considered as the true (baseline) values are compared with the FFNN and LSTM predicted solutions. In Fig. \ref{fig: CD-LSTM}, the evolution of the extracted latent spaces $\mathbf{\tilde{z}}$ for the testing parameter $\bm{\mu}^{te}=(104, \: 167)$ m/s is presented. We observe that the FFNN can accurately predict the first $w$ components ($w=10$) of the latent space time sequence, corresponding to the time interval $t$=[0,$\:$0.001] s, required by the LSTM, during the online phase to predict the entire evolution in the time. Furthermore, in Fig. \ref{fig: CD-LSTM} it is evident that the LSTM  predicts precisely the dynamic behavior until $t$=0.0075 s.
\begin{table}[t]
\caption{NN-configs. and efficiency of the DL-models in terms of accuracy for the CD test case. The GPU computational times of the ROM deployment are presented against the HFM simulation time that requires 15 m for a given parameter $\bm{\mu}_i$.}
\centering{
\begin{tabular}{|c|c|c|c|}
\hline
\diagbox[width=\dimexpr \textwidth/8+2\tabcolsep\relax, height=1.5cm]{$\textbf{NN-configs}$}{$\textbf{DL-model}$}       
& $\textbf{CAE}$ & $\textbf{FFNN}$ & $\textbf{LSTM}$ \\ \hline
$$\textbf{learning rate}$$ & $0.001$        & $0.2$           & $0.001$         \\ \hline
$\textbf{batch}$         & $8$           & $1$          & $5$             \\ \hline
$\textbf{epochs}$        & $1500$         & $1000$          & $1500$             \\ \hline
$\textbf{val. loss}$  & $5.6\times10^{-6}$ & $5.8\times10^{-5}$ & $2.7\times10^{-5}$ \\ \hline
$\textbf{train. loss}$ & $3.8\times10^{-5}$ & $7.3\times10^{-5}$ & $1.7\times10^{-4}$ \\ \hline
$\textbf{training time}$ & $40$ m         & $15$ m          & $50$ m          \\ \hline
$\textbf{testing time}$  & $0.08$ s       & $0.015$ s       & $0.1$ s         \\ \hline
$\textbf{speed up}$   & $1.1\times10^{4}$  & $6\times10^{5}$    & $9\times10^{4}$    \\ \hline
\end{tabular}}\label{Table: CD}
\end{table} 

\begin{figure}[H]
\centering\includegraphics[width=0.98\linewidth]{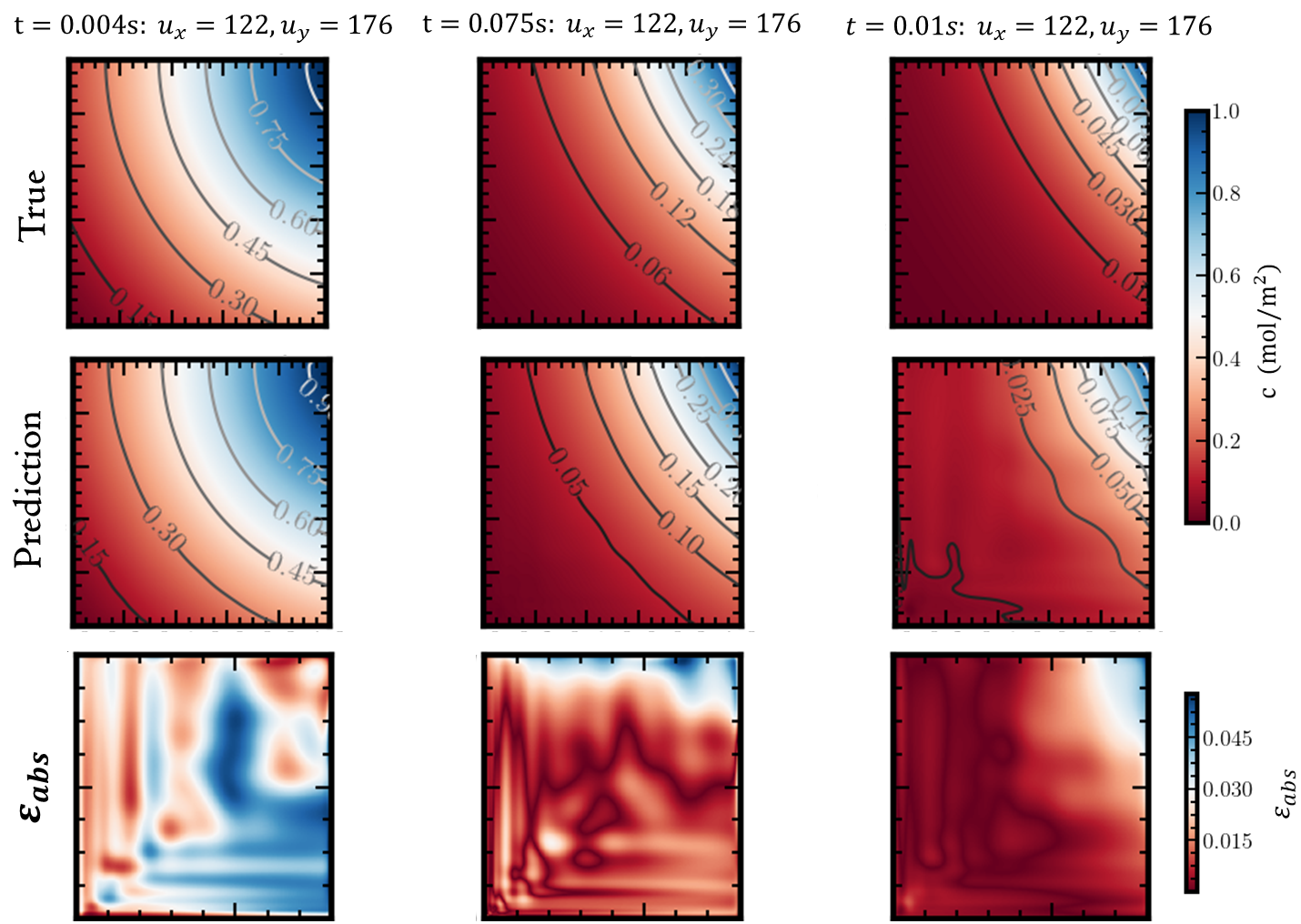}
\caption{Comparison between the true and predicted solution at $t=0.004s$ (left) and $t=0.0075s$ (middle) for the testing set $\bm{\mu}^{te}=(122,176)$ m/s. The ability of the $\textit{FastSVD-ML-ROM}$ to forecast is presented at $t=0.01$ s (right).} \label{FOM-CD}
\end{figure}

In Fig. \ref{FOM-CD}, the results of the true and predicted concentration field for the test set $\bm{\mu}^{te}=(122, \: 176)$ m/s are compared. We note the ability of the \textit{FastSVD-ML-ROM} to predict the concentration field for time $t\leq0.0075$ s, and forecast for time $t>0.0075$ s. The predicted solution field at $t=0.004$ s and $t=0.0075$ s, is in good agreement with the true solution. To test the ability of the developed surrogate model to forecast in unseen times, the solutions of the CD equation until $t=0.01$ s are presented. 
We observe that the ROM can extrapolate in time by an additional $25\%$ of the training time, providing good solutions with a low-absolute error at the entire domain, and higher error values at the corner of the plate. Moreover, to gain a better insight in the computational accuracy of ROM, we calculate the $\varepsilon_{rmse}$ error for the predicted solutions. These errors are: $\varepsilon_{nrms}$ = $4\times10^{-3}$ at $t=0.004$ s, $\varepsilon_{nrms}$ = $3.7\times10^{-3}$ at $t=0.0075$ s while the $rsme$ error is $\varepsilon_{nrms}$ = $9\times10^{-2}$ at $t=0.01$ s. The absolute error slightly increases for later times since the surrogate model cannot fully capture the field at the upper right corner.

\begin{figure}[t]
\centering\includegraphics[width=0.92\linewidth]{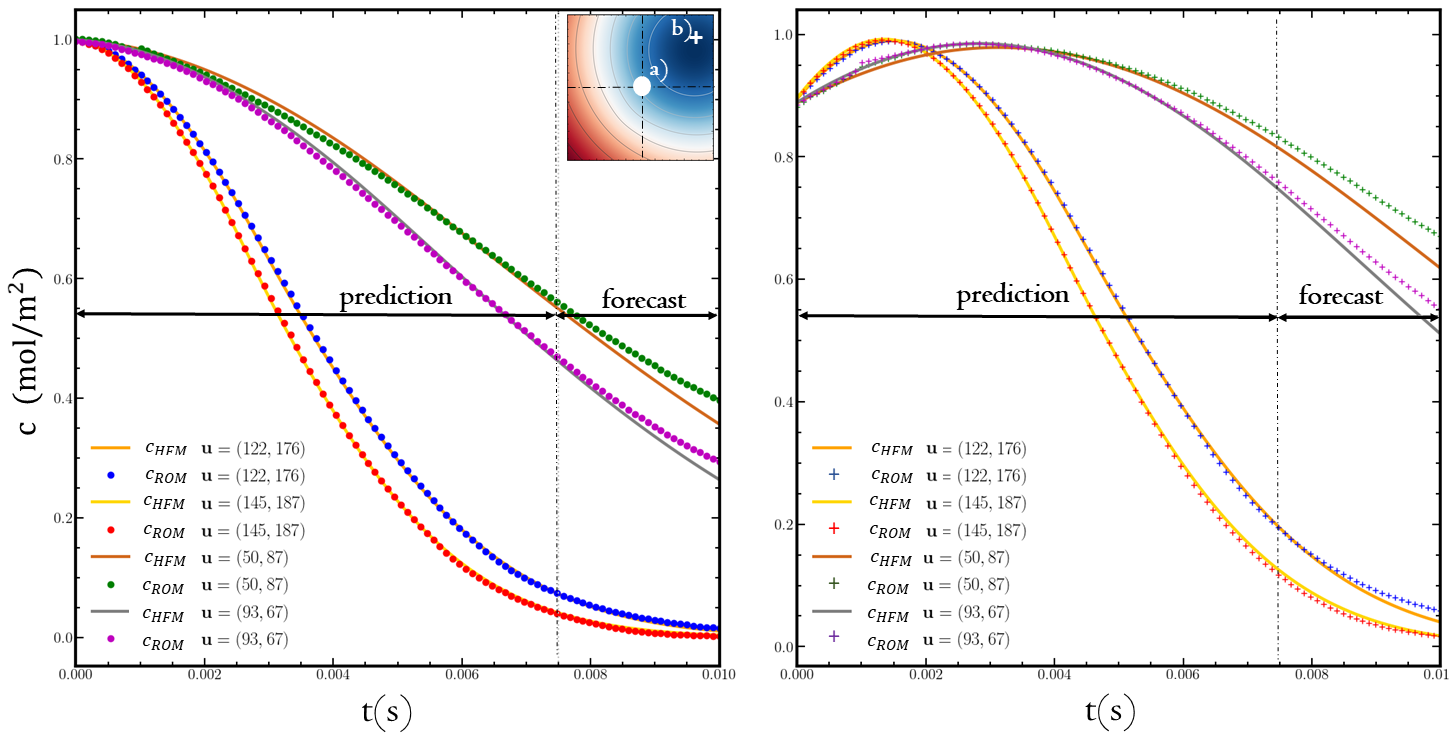}
\caption{Monitoring of the true and predicted solutions on point a ($\bullet$) located at the central of the domain (left) and on point b (+) near the right corner of the domain (right) for $\bm{\mu}^{te}=\{ (122,176), (145, 187), (50, 87), (93, 67)\}$  m/s.
} \label{1D-plot:FOM/ROM}
\end{figure}

In Fig. \ref{1D-plot:FOM/ROM}, we showcase the efficiency of the developed framework to monitor the evolution of the concentration along with time on the central node (a) (Fig. \ref{1D-plot:FOM/ROM},a), and on a node located near the right corner (b) (Fig. \ref{1D-plot:FOM/ROM},b), for the parameter set: $\bm{\mu}^{te}=\{(122, 176), \: (145, 187),\: (50, 87),\: (93, 67)\}$ m/s. It is observed that the trend of the function can be accurately captured both during the prediction and forecast phase, for higher and lower velocity values.

\begin{figure}[H]
\centering\includegraphics[width=0.93\linewidth]{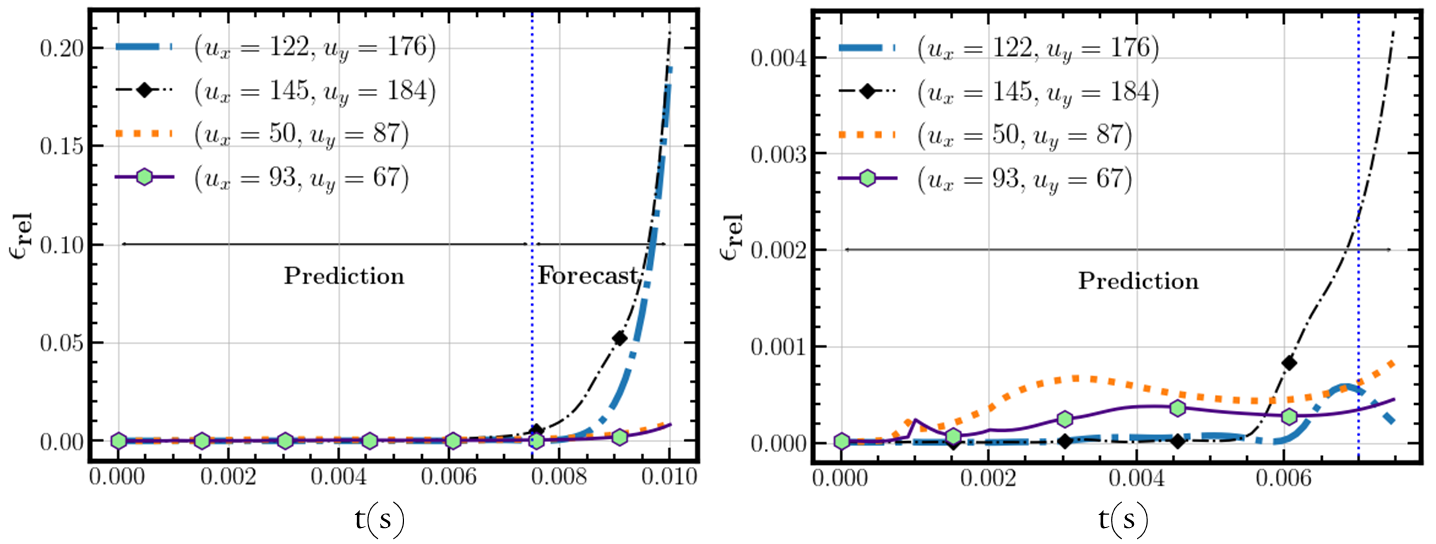}
\caption{Evolution of the $\varepsilon_{rel}$ for predicted and forecast values versus time for the testing parameters \: \: \: \: \: \: \: $\bm{\mu}^{te}=\{(122,176), \: (145, 187), \: (50, 87), \:  (93, 67)\} \: m/s $.}\label{1__rel.png}
\end{figure}

We finally access the efficiency and accuracy of the proposed ROM, with respect to the $\varepsilon_{rel}$ error for $\bm{\mu}^{te}=\{(122,176), (145, 187), (50, 87), (93, 67)\} \: m/s $, presented in Fig. \ref{1__rel.png}. As observed, during the prediction phase the error indicator has a peak value equal to $\varepsilon_{rel}=0.004$ and during the forecast phase, the error is increased for larger times, reaching approximately a value close of $\varepsilon_{rel}\approx0.2$ at $t=0.01s$.
\vspace{-5pt}

\subsection{Benchmark problem 2: Flow around a cylinder} \label{Cylinder}
In this example, the unbounded external flow around a cylinder with circular cross-section is examined.  We numerically solve the incompressible Navier-Stokes (N-S) equations---in their primitive variables (velocity \(\bm{u}\) and pressure \(p\)) formulation--- written as:
\begin{equation} \label{eq:41}
     \left(
        \frac{\partial\bm{u}}{\partial t}
        + \bm{u} \cdot \bm{\nabla}\bm{u}
    \right) = 
    - \bm{\nabla} p
    + \bm{\nabla} \cdot 2v\bm{\varepsilon}(\bm{u}) 
    + \bm{f},
\end{equation}
\begin{equation} \label{eq:42}
    \nabla \cdot \bm{u}  = 0,
\end{equation}
The term \(\bm{\varepsilon}(\bm{u} )\) is the strain-rate tensor
defined as:
\begin{equation} \label{eq:43}
    \bm{\varepsilon}(\bm{u}) = \frac{1}{2}\left(
        \bm{\nabla}\bm{u} + \left( \bm{\nabla}\bm{u} \right)^\mathrm{T}
    \right).
\end{equation}
while $\bm{f}$ are the body forces and \(v\) is the kinematic viscosity of the fluid.

\begin{figure}[h]
\centering\includegraphics[width=0.95\linewidth]{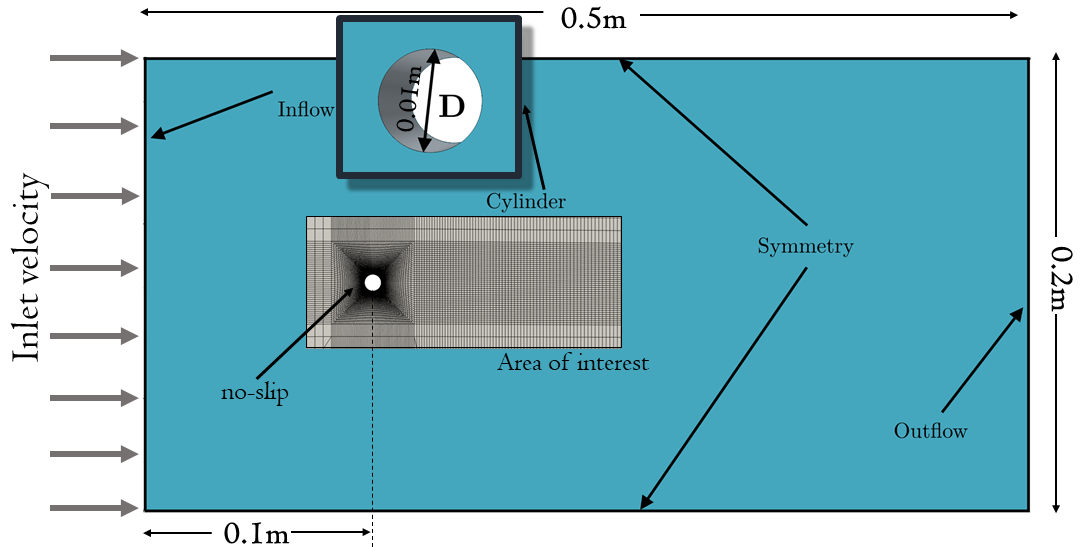}
\caption{The geometry and the boundary conditions of the flow around cylinder case.} \label{geom:flow}
\end{figure}

The flow domain is large compared to the dimensions of the cylinder as shown in Fig. \ref{geom:flow}. The center of the 0.01m in diameter cylinder, is located at point \( O (0.1m, 0.1m)\) of a square domain with dimensions $0.0 \leq x \leq 0.5$ and $0.0 \leq y \leq 0.2$. The flow is not perturbed near the inlet, which is located far from the cylinder. The inflow is defined from the left side and the outflow from the right side, respectively. No-slip conditions are defined on the wall and the cylinder. The flow domain is discretized with 211,000 finite volume cells, locally refined in the vicinity of the cylinder. The grid resolution ensures a grid independent numerical solution. We numerically solve the governing equations using the commercial software Siemens Star CCM+ \cite{guide2020starccm+}. The fluid's density is $\varrho=997 \: kg/m^3$ and the kinematic viscosity is $v=1.0034 \: mm^2/s$. 

The main input parameter that formulates the snapshot matrix is the the Reynolds number (Re). We keep the indicated region that consists the area of interest, including 13,834 cells (Fig. \ref{geom:flow}). We consider the Reynolds number as the parameter, in the space \(P = [100, \: 210]\). This range is chosen to generate non-linear flow patterns, e.g. karman vortex streets behind the cylinder. We uniformly sample $m=9$, training $\bm{\mu}^{tr}$ and $n=2$ testing $\bm{\mu}^{te}$ parameters, as well. The unsteady case is solved with the non-iterative PISO algorithm, over the interval \( t = [0, \: 120]\) s, with a time-step $dt=5\times 10^{-4}$ s resulting in 240,000 snapshots. We then choose 480 solutions, with a step equal to 500, and we keep the last 260 instances, belonging to the time range \( t = [55, \: 120]\) s. We implement the $\textit{FastSVD-ML-ROM}$, to reconstruct the velocity magnitude field in real-time, during the online phase, for a given parameter set $\bm{\mu}^{tr}$.

\begin{figure}[t]
\centering\includegraphics[width=1\linewidth]{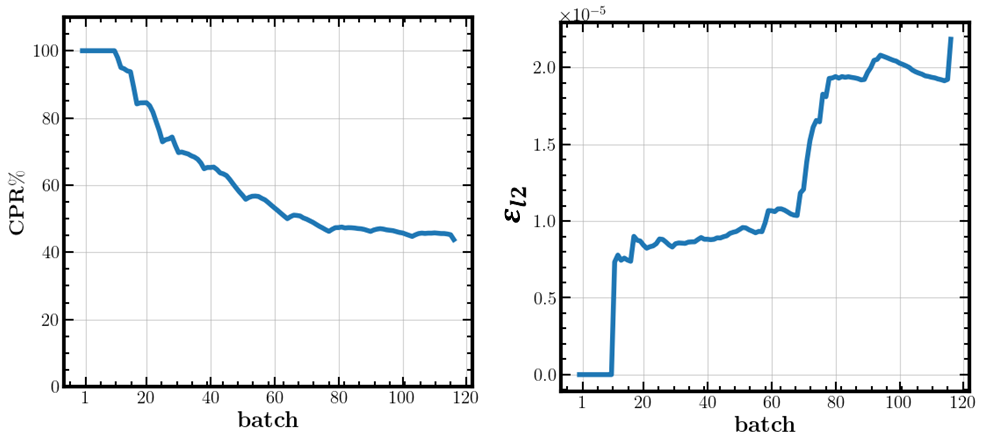}
\caption{The implementation of the truncated SVD update for the flow around a cylinder test case. The evolution of the compression ration (CPR) (left) and the $\epsilon_{l2}$ (right) are presented versus the batches using a truncation error $\epsilon = 7.6 \times 10^{-7}$.}\label{SVD art}
\end{figure}

\begin{figure}[t]
\centering{
\includegraphics[width=1\linewidth]{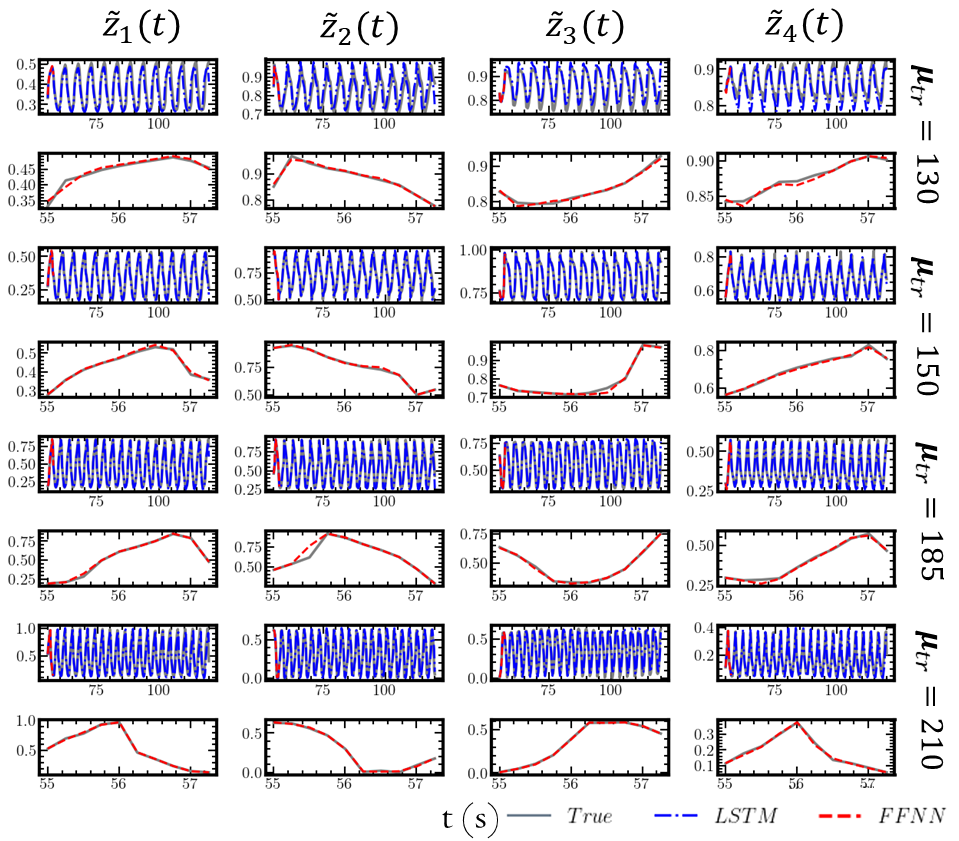}
\caption{True latent variables extracted by the trained decoder, LSTM and FFNN predictions (top) at \( t = [55,  120] \: s\), and FFNN predictions (bottom) at \(t = [55, 65]\) s for each training parameter $\bm{\mu}^{te}$ = \{130, 150, 185, 210\}.}\label{2__LSTM_FFNN.png}}
\end{figure}

Regarding the truncated SVD update, we define \( l = 13\) subsets, resulting in \( s = 20\) time solutions for each subset. The accuracy is set to $\varepsilon=7.6 \times 10^{-6}$, leading to \( n = 1024\), linear basis vectors. In Fig. \ref{SVD art}, it is shown that the snapshot matrix can not be compressed for the first 13 batches. This indicates that the linear dimensionality reduction technique can not identify a reduced basis for this part of the snapshot matrix. After the first 12 batches, it is observed that the decay of the CPR is approximately exponential, while the maximum compression achieved is about 40$\%$.
Regarding the CAE, we specify the latent space dimension to be $q=4$, resulting to a compression of 0.39$\%$ with respect to the linear projected data. For the LSTM network, a time window \( w = 10\) is pre-defined. The NN-configurations for each DL-model are presented in Table \ref{Table: Cylinder} in detail and the NN architectures are presented in Appendix \ref{NN arch}. As concerns the implementation of the NAS (Appendix \ref{NAS}), the sigmoid activation function is utilised, presenting the minimum validation and training loss.
\begin{table}[H]
\caption{NN-configurations and efficiency of the DL-models in terms of accuracy for the flow around a cylinder test case. The GPU computational times of the ROM deployment are presented against the HFM simulation time that requires 12 h for a given parameter $\bm{\mu}_i$.}
\centering{
\begin{tabular}{|c|c|c|c|}
\hline
\diagbox[width=\dimexpr \textwidth/8+2\tabcolsep\relax, height=1.5cm]{$\textbf{NN-configs}$}{$\textbf{DL-model}$}             & $\textbf{CAE}$ & $\textbf{FFNN}$ & $\textbf{LSTM}$ \\ \hline
$$\textbf{learning rate}$$ & $0.0005$        & $0.1$           & $0.0001$         \\ \hline
$\textbf{batch}$         & $20 $           & $50$          & $10$             \\ \hline
$\textbf{epochs}$        & $2000$         & $50000$          & $1500$                 \\ \hline
$\textbf{val. loss}$  & $4.7 \times 10^{-6}$ & $3.1 \times 10^{-5}$ & $2 \times 10^{-5}$ \\ \hline
$\textbf{train. loss}$ & $5 \times 10^{-5}$ & $6.4 \times 10^{-5}$ & $1.7  \times 10^{-4}$ \\ \hline
$\textbf{training time}$ & $35$ m         & $25$ m          & $45$ m          \\ \hline
$\textbf{testing time}$  & $0.09$ s       & $0.011$ s       & $0.2$ s         \\ \hline
$\textbf{speed up}$   & $4.8 \times 10^{6}$  & $3.9  \times 10^{8}$    & $2.1 \times 10^{6}$    \\ \hline
\end{tabular}}\label{Table: Cylinder}
\end{table}
\vspace{-0.4cm}
In Fig. \ref{2__LSTM_FFNN.png}, to access the accuracy of the FFNN and the LSTM predictions, we present the evolution of the latent space representations along with time for various training parameter sets $\bm{\mu}^{tr}=\{ 130, 150, 185, 210 \}$. It is observed that the FFNN can precisely capture the first sequential data containing the latent space representations until \(t = 65\) s. Moreover, the LSTM can accurately predict the time evolution of the latent space representations for the entire time range (\(t = [120, 210] \:s)\). It is evident that the combination of the FFNN and the LSTM can capture the dynamics of the latent space representations extracted of the trained encoder, part of the CAE.

\begin{figure}[t]
\centering\includegraphics[width=1\linewidth]{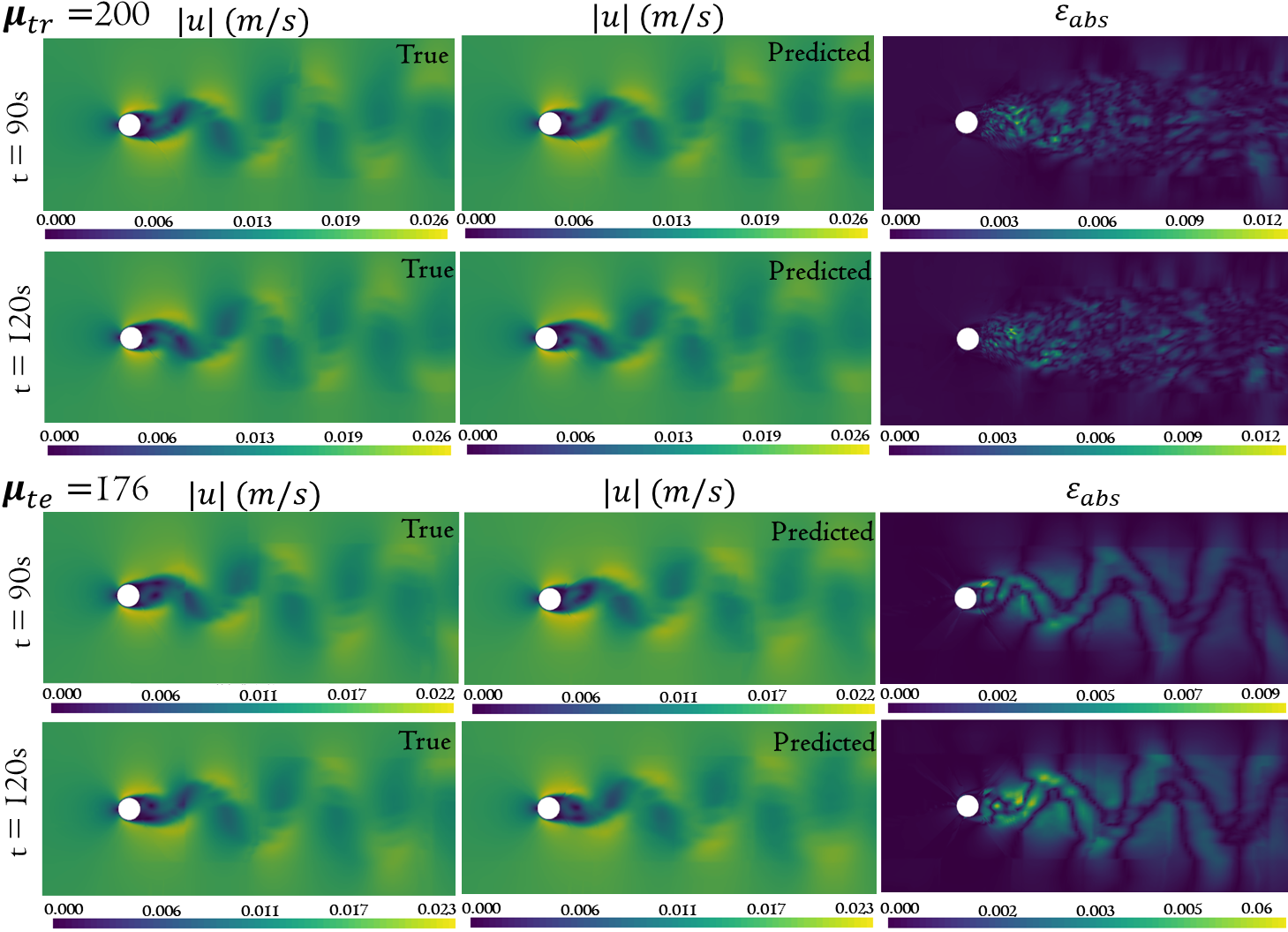}
\caption{Comparison of the true and predicted velocity field for $\bm{\mu}^{te}=200$ (top) at \(t = \)90 s and \(t = \)120 s with a $\varepsilon_{rmse}$ equal to $3.9 \times 10^{-3}$ and $3.4 \times 10^{-3}$. \label{2__FOM_ROM1.png}}
\end{figure}

The accuracy of the proposed methodology for the flow around a cylinder case is presented in Fig. \ref{2__FOM_ROM1.png}. It is shown that the figures of the predicted velocity field are in good agreement with the velocity field obtained by the HFM and the ROM, can capture the vortices along the cylinder wake. The periodic pattern of the vortex shedding is clearly observed both for the testing $\bm{\mu}^{te}=200$ and training $\bm{\mu}^{te}=176$ parameters. 

\begin{figure}[H]
\centering\includegraphics[width=1\linewidth]{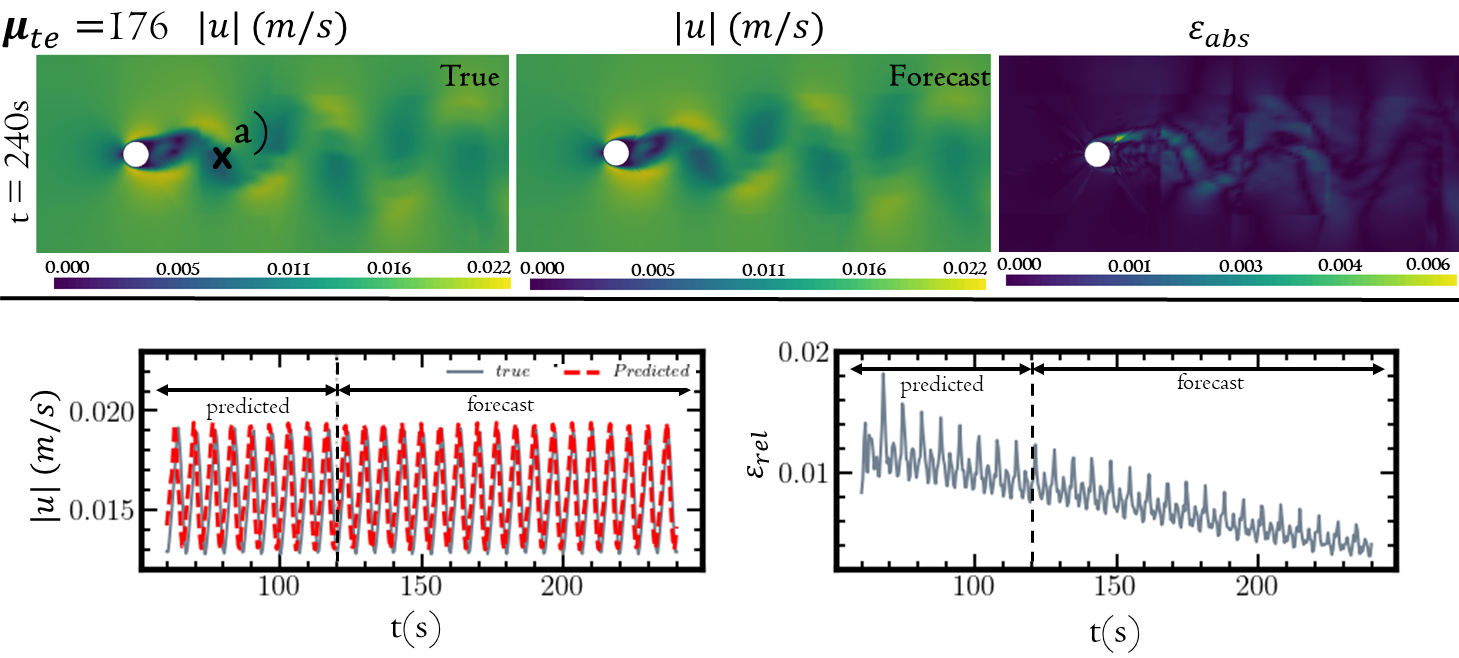}
\caption{Comparison of the true and forecast velocity at \(t = 240\) s, for the testing parameter $\bm{\mu}^{te}$ = 176 (top), time tracing on point $a$ (x), behind the cylinder (bottom-left), and $e_{rel}$ versus time (bottom-right).\label{2__FOM_ROM_fore.png}}
\end{figure}

In Fig. \ref{2__FOM_ROM1.png}, larger absolute values are obtained in the area behind and close to the cylinder, where the von-karman vortices are starting to form. In Fig. \ref{2__FOM_ROM_fore.png}, we demonstrate the ability of the surrogate model to forecast, in unseen times, \(t > 120\) s. The velocity magnitude on the point $a$ behind the cylinder and the $\varepsilon_{rel}$ along with time are depicted in Fig. \ref{2__FOM_ROM_fore.png} (bottom-left) plot while the evolution of the relative error $e_{rel}$ along with time is presented in n Fig. \ref{2__FOM_ROM_fore.png} (bottom-right). It is shown that the LSTM network can accurately capture the dynamics, enabling an additional 100$\%$ forecast, starting from the predicted solutions with a decreased relative error $e_{rel}$. 

\subsection{Benchmark problem 3: Blood flow in an arterial segment} \label{Arterial}
\begin{figure}[ht]
\centering\includegraphics[width=1\linewidth]{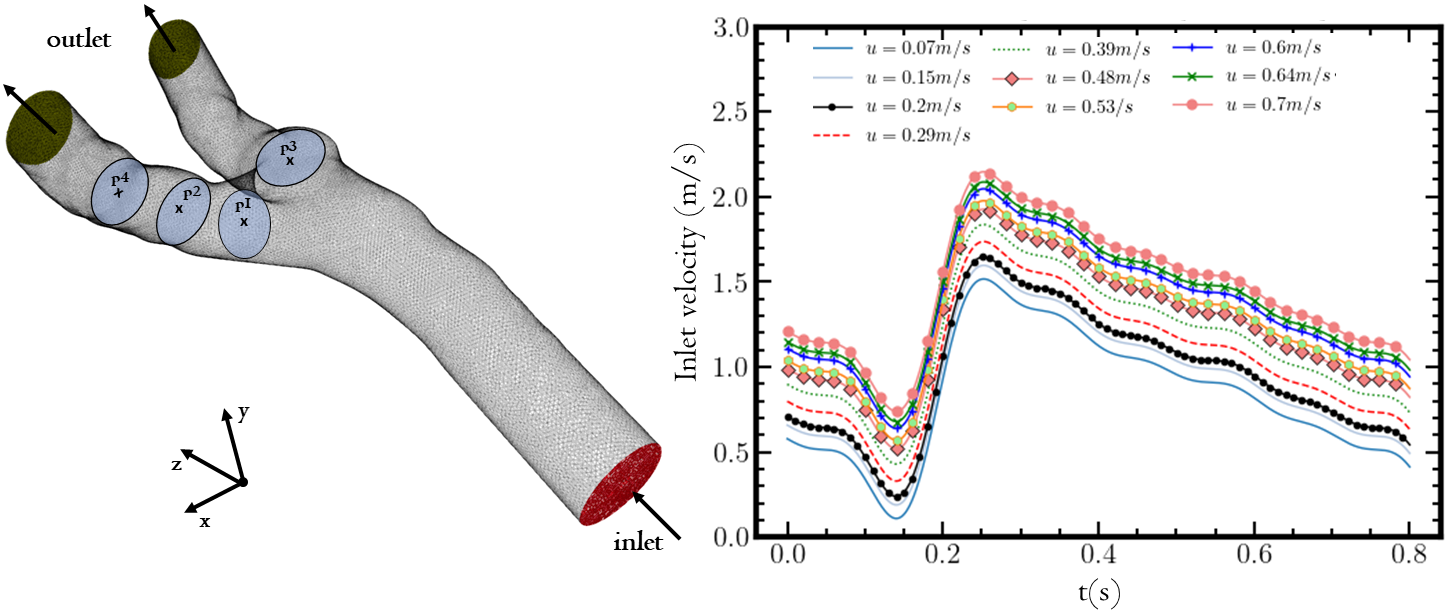}
\caption{ Geometry of the arterial lumen including  the boundary condition (left) and the inlet velocity during a heart cycle on the central node of the inlet face. \label{3__geometry.png}}
\end{figure}

In the final example, the blood flow in a bifurcation is simulated (shown in. Fig. \ref{3__geometry.png}) by utilizing the the Navier-Stokes equations, presented in Eq. \ref{eq:41}, \ref{eq:42} and \ref{eq:43}. The inlet and two outlets of the flow domain were extruded to ensure that the flow is fully developed.
The flow domain is discretised using tetrahedral elements. We set the total time for the simulation to \(T = 0.8\) s (one cardiac cycle), and the time step is equal to \(dt = 1 \times 10^{-3}\) s, resulting to 800 time solutions. At the inlet, the pulsatile velocity waveform is applied, a parabolic velocity profile since the womersley number is small, and zero pressure boundary conditions at the two outlets, as shown in Fig. \ref{3__geometry.png}. At the remaining walls, we apply no-slip boundary conditions. The Newtonian model for blood flow is employed, with kinematic viscosity $v$ = \( 3.26\  mm^2/s\) and density $\varrho$ = \(1056\ kg/m^3\). The governing equations are solved using the incremental pressure correction scheme, the using FENiCS FEM solver \cite{logg2012automated}. 

In this test case, we aim to reconstruct the velocity field components $\bm{u}_{x}, \bm{u}_{y}$ and $\bm{u}_{z}$ for a given inlet velocity. Therefore, the governing equations are solved for $m = 10$ training and $n = 2$ testing parameters respectively, in the range \(\bm{\mu} = [0.07 - 0.7]\) m/s. Given the time range and the time step, 800 time solutions are generated for each hihg fidelity solution. To reduce the computational time needed to train the DL-models, instead of using the entire data, we define a step equal to 5, and thus $N_t=160$ solutions are utilised for each parameter. 

\begin{figure}[t]
\centering\includegraphics[width=1\linewidth]{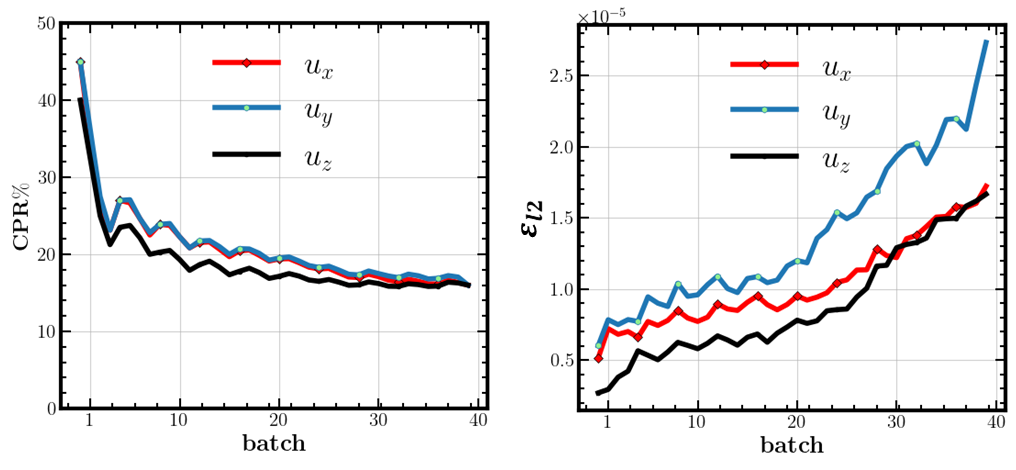}
\caption{The implementation of the truncated SVD update for the CD test case. The evolution of the compression ration (CPR) (left) and the $\epsilon_{l2}$ (right) are presented versus the batches using a truncation error $\epsilon = 8.4 \times 10^{-6}$ for the x component,  $\epsilon = 9.8 \times 10^{-6}$ for the y component and $\epsilon = 6.2 \times 10^{-6}$ for the  z component, as well.  \label{SVD res}}
\end{figure}

In Table \ref{Table: Artery} the DL-configurations, are presented in detail, while the NN architectures are listed in Appendix \ref{NN arch}. Regarding, the implementation of the NAS (Appendix \ref{NAS}), the ReLU activation function is utilised, presenting the minimum validation and training loss. As concerns the SVD update, each HFM simulation is divided to \(s = 4\) subsets, resulting in \(l = 40\) time solutions. We define a separate reduced basis for each velocity set with the same dimension. This is achieved by appropriately specifying the truncation error for the SVD update algorithm. In particular, we set \( \varepsilon = 8.4 \times 10^{-6} \) for the $x$-axis component, \(\varepsilon = 9.8 \times 10^{-6}\) for the $y$-axis component and \(\varepsilon = 6.2 \times 10^{-6}\) for the $z$-axis component respectively, resulting to 256 basis vectors for each component. In Fig. \ref{SVD res}, it is shown that the CPR for each velocity component presents a similar behavior: the evolution along with time, exponentially decreases, until it reaches a plateau close to 18 \%. The $\varepsilon_{l2}$ error obtains its larger value for the $\bm{u}_y$ component, since the basis $\bm{u}_y$ is computed with a larger value of $\varepsilon$. Rereading the DL-models, a latent dimension \(q = 4\) is defined for the CAE leading to a compression 1.56 \% with respect to the linear projected data for each velocity component. Concerning the training of the LSTN network, a time window \(w = 30\) is predetermined. 

\begin{figure}[H]
\centering\includegraphics[width=0.92\linewidth]{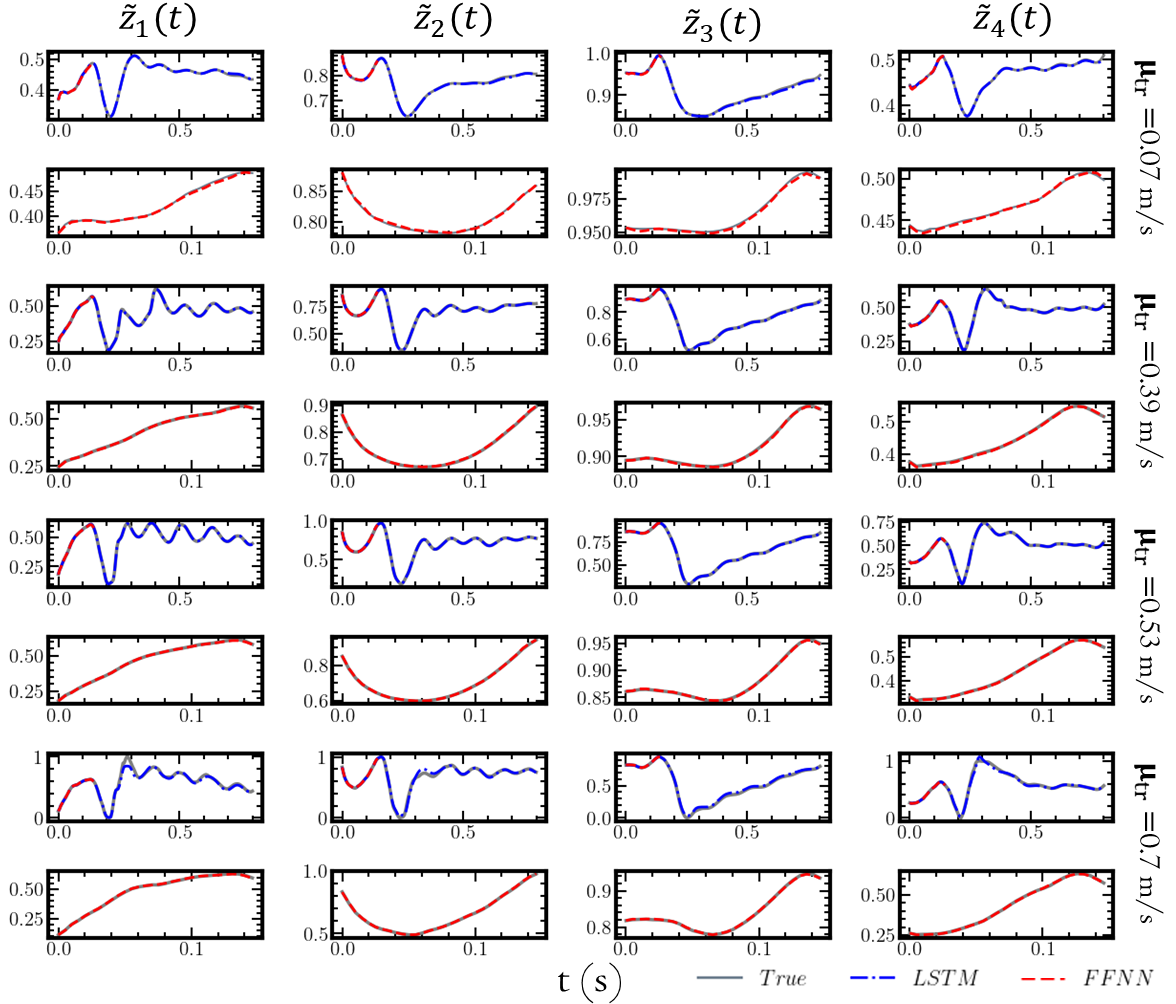}
\caption{True latent variables extracted by the trained decoder, the LSTM and the FFNN predictions (top) at $t$=[0, 0.8] s, and the FFNN predictions (bottom) at $t$=[0, 0.15] s for the training parameter $\bm{\mu}^{tr}=\{0.07, \: 0.39, \: 0.53, \:0.7\}$ m/s.\label{3__LSTM_FFNN.png}} 
\end{figure}

In Fig. \ref{3__LSTM_FFNN.png}, we present the ability of the $\textit{FastSVD-ML-ROM}$ framework, to predict the dynamics along with time, for the testing parameters $\bm{\mu}^{te}=\{0.07, 0.39, 0.53, 0.7\}$ m/s. 


It is shown that both the LSTM network and the FFNN can accurately predict the evolution of the latent space representations, with each variable presenting a different dynamical behavior. Regarding the top plots for each parameter, we observe that after \(t > 0.5\) s, the curve becomes more oscillatory as the inlet velocity increases, and the time prediction models precisely capture the non-linear evolution in time. 
\begin{table}
\caption{NN-configs. and efficiency of the NN-models in terms of accuracy for the flow in an arterial segment test case. The GPU computational times of the ROM deployment are presented against the HFM simulation time that requires 8 h for a given parameter $\bm{\mu}_i$.}
\centering{
\begin{tabular}{|c|c|c|c|}
\hline
\diagbox[width=\dimexpr \textwidth/8+2\tabcolsep\relax, height=1.5cm]{$\textbf{NN-configs}$}{$\textbf{DL-model}$}             & $\textbf{CAE}$ & $\textbf{FFNN}$ & $\textbf{LSTM}$ \\ \hline
$$\textbf{learning rate}$$ & $0.0005$        & $0.01$           & $0.0001$         \\ \hline
$\textbf{batch}$         & $20 $           & $4$          & $10$             \\ \hline
$\textbf{epochs}$        & $2000$         & $10000$          & $35000$          \\ \hline
$\textbf{val. loss}$  & $7.4\times10^{-5}$ & $3.9\times10^{-6}$ & $1.1\times10^{-5}$ \\ \hline
$\textbf{train. loss}$ & $9.2\times10^{-4}$ & $5.4\times10^{-6}$ & $3\times10^{-4}$ \\ \hline
$\textbf{training time}$ & $2$ h         & $30$ m          & $3$ h          \\ \hline
$\textbf{testing time}$  & $0.05$ s       & $0.013$ s       & $0.2$ s         \\ \hline
$\textbf{speed up}$   & $5.8\times10^{5}$  & $6\times10^{6}$    & $1.4\times10^{5}$    \\ \hline
\end{tabular}}\label{Table: Artery}
\end{table}

\begin{figure}[h] 
\centering\includegraphics[width=0.9\linewidth]{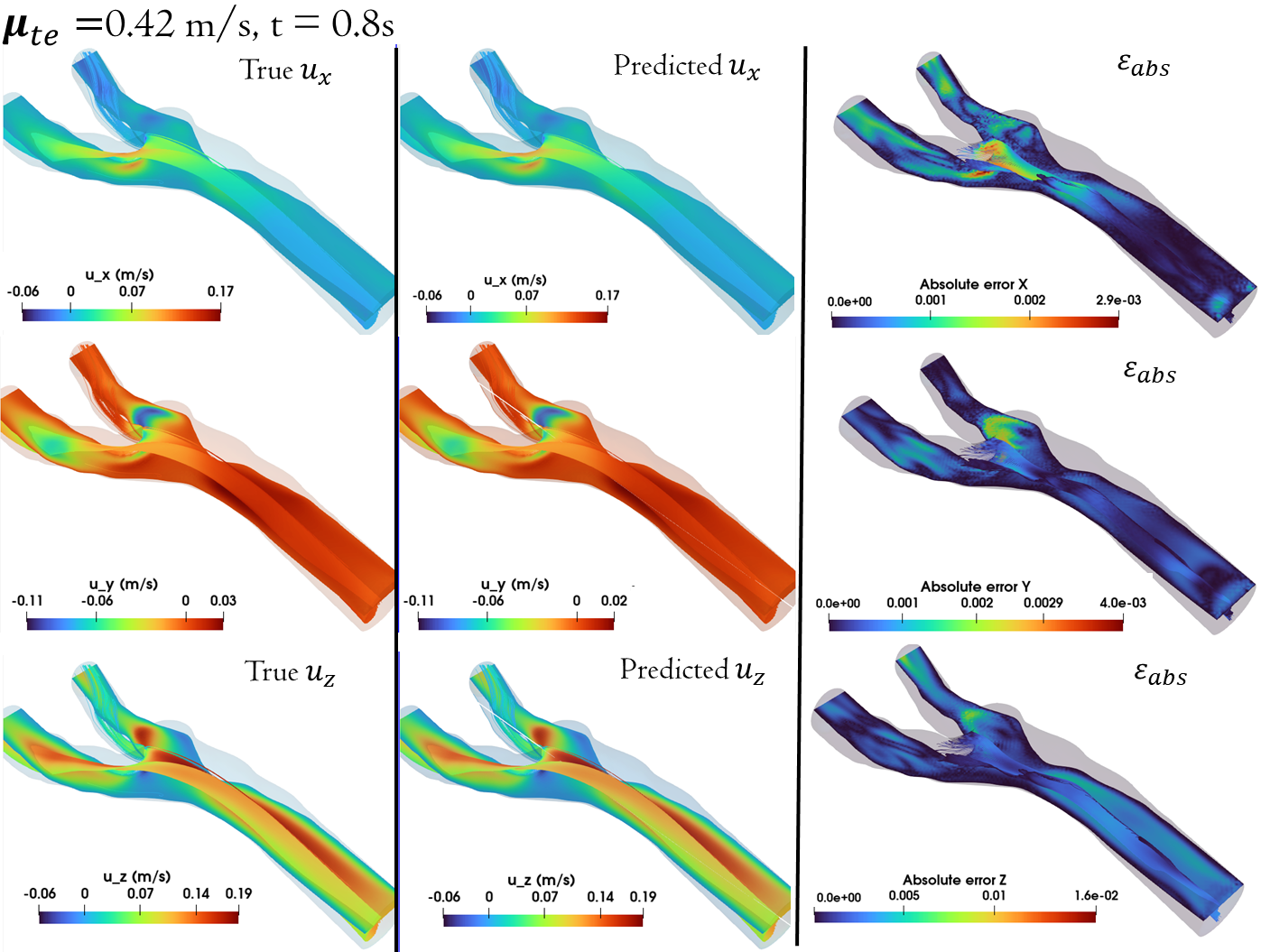}
\caption{Comparison of the true and predicted velocity components for $\bm{\mu}_{te}=0.42$ m/s during $t=0.8$ s at $x$-axis (left), $y$-axis (middle) and $z$-axis (right). \label{3__fomrom042.png}}
\end{figure}

\begin{figure}[H]
\centering\includegraphics[width=0.9\linewidth]{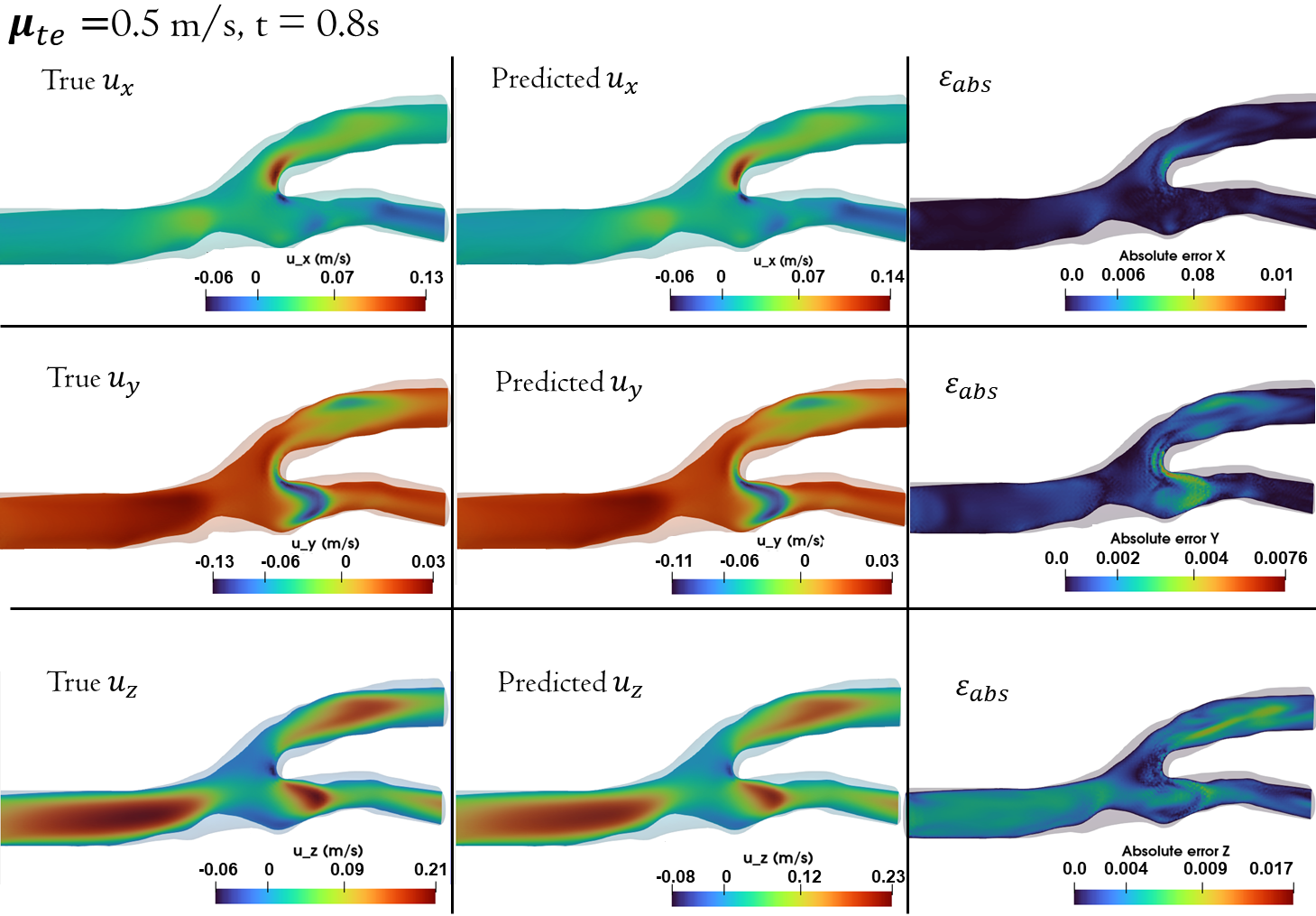}
\caption{Comparison of the true and predicted velocity components for $\bm{\mu}_{te}=0.5$ m/s during $t=0.8$ s at $x$-axis (left), $y$-axis (middle) and $z$-axis (right). \label{3__fomrom05.png}}
\end{figure}

To assess the performance of the $\textit{FastSVD-ML-ROM}$, in Fig. \ref{3__fomrom042.png} and \ref{3__fomrom05.png}, we present for $\bm{\mu}^{te}=\{0.42, 0.5\}$ m/s, the predicted and true solutions for $\bm{u}_x$, $\bm{u}_y$, and $\bm{u}_z$. As it is evident, the velocity reconstruction in both the x, y, and z-axis are in excellent agreement with the HFM (true) solutions. 
The $\varepsilon_{abs}$ obtains its largest value in in $z$-axis compared to the $x$-axis and $y$-axis for both parameters. For the $\bm{\mu}^{te} = 0.42$ m/s, the $\bm{u}_x$ component presents \(\varepsilon_{rmse} = 3.6 \times 10^{-3}\), the $\bm{u}_y$ component \( \varepsilon_{rmse} = 3.9 \times 10^{-3}\) and the $\bm{u}_z$ component \( \varepsilon_{rmse} = 9.6 \times 10^{-3}\), respectively. Regarding the $\bm{\mu}^{te} = 0.5$ m/s, \( \varepsilon_{rmse} = 3.5 \times 10^{-3}\) is obtained for $\bm{u}_x$ component, \(\varepsilon_{rmse} = 5.6 \times 10^{-3}\) for $\bm{u}_y$ component and \(\varepsilon_{rmse} = 1.3 \times 10^{-2}\) for $\bm{u}_z$ component.

\begin{figure}[t]
\centering\includegraphics[width=1\linewidth]{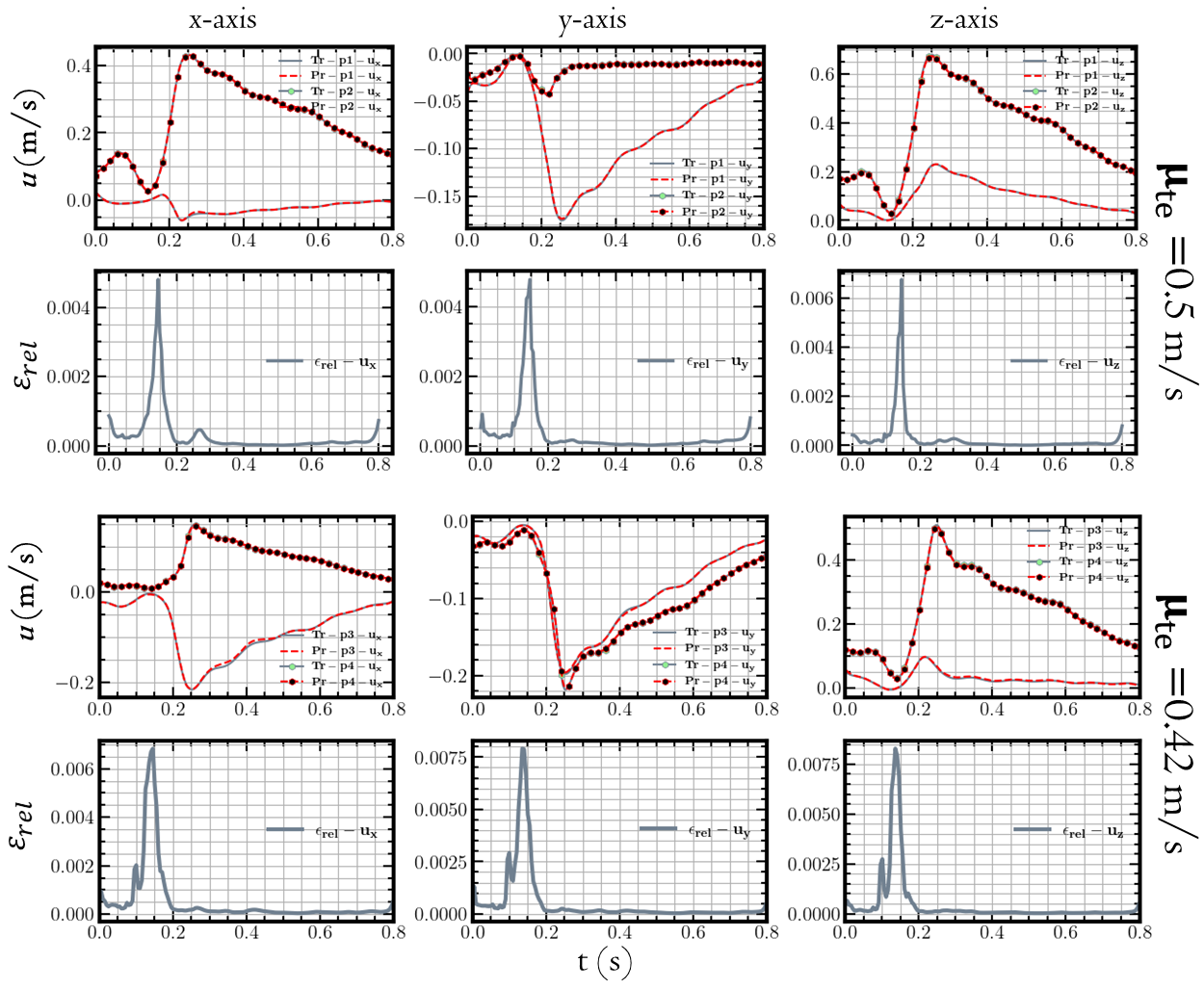}
\caption{Velocity along with time for points p1,p2,p3 and p4 (shown in Fig. \ref{geom:flow}) and \(\varepsilon_{rel}\) errors versus time for $\bm{\mu} = $0.5 m/s and $\bm{\mu}_{te } = $0.42 m/s. \label{3__1Dplot.png} }
\end{figure}

\begin{figure}[H]
\centering\includegraphics[width=0.9\linewidth]{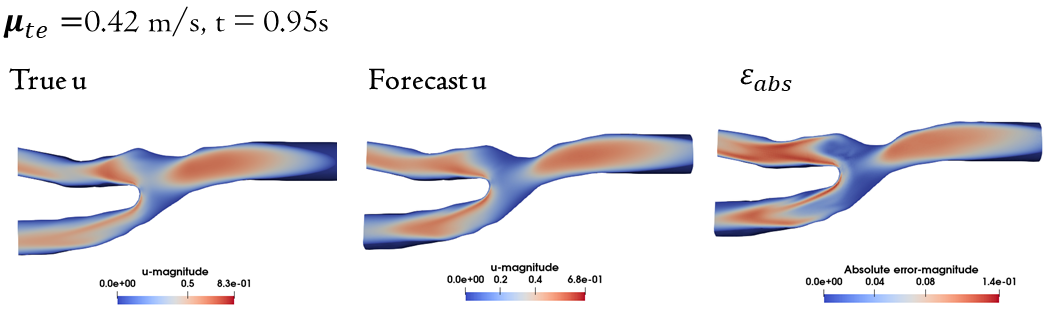}
\caption{Comparison of the true and forecast velocity magnitude solutions for \(\bm{\mu} = 0.42 \) m/s. \label{3__forecast.png} }
\end{figure}

In Fig. \ref{3__forecast.png}, we present the true, forecast and $\varepsilon_{abs}$ error for \( \bm{\mu}_{te} = 0.42 \) m/s, at time \(t = 0.95\) s. In Fig. \ref{3__1Dplot.png}, we monitor the velocity on points p1, p2, p3 and p4  located close to the centroid of the arterial segment at the indicated locations (Fig. \ref{3__geometry.png}). 
We demonstrate the ability of the \textit{FastSVD-ML-ROM}, to accurately predict the evolution of the velocity components at various locations near the bifurcation region. The $\varepsilon_{rel}$ is increased both for $\bm{u}_x$, $\bm{u}_y$ and $\bm{u}_z$ at \(t = 0.2 \) s since the inlet velocity is increased.
\setlength{\parskip}{0pt}

\section{Conclusions} \label{Conclusions}
In the present work, a non-intrusive ROM platform has been proposed, mentioned as \textit{FastSVD-ML-ROM}, to provide fast and accurate solutions for transient, (non)linear physics-based simulation in real-time. The developed surrogate model targets on building the virtual replica of the physical asset and can be easily adapted in a digital twin system to reconstruct the field of interest in a limited amount of time. The \textit{FastSVD-ML-ROM} framework includes a truncated SVD update methodology, to identify a linear basis of large-scale numerical models and reduce the dimensions of the computed simulations. CAE is utilized to identify latent variables of the projected data through non-linear dimensionality reduction. LSTM and FFNN networks are implemented to predict and forecast the solutions. 

To illustrate the robustness of \textit{FastSVD-ML-ROM} and the speed-up compared to the computational time of the numerical solvers, we present three benchmark problems dealing with, \textit{i)} the 2D advection-diffusion equation, \textit{ii)} the 2D flow around a cylinder and \textit{iii)} the 3D blood flow in an arterial segment. The comparison between the high fidelity solutions and the ROM reconstructions demonstrates the ability of the FastSVD-ML-ROM to accurately predict the unsteady fields of interest, for a given parameter vector in real-time. It is also evident that the combination of the LSTM and the FFNN networks, assisted by the NAS technology, can efficiently handle (non)linear dynamics while extrapolating in time. Regarding the linear PDE, the ROM can provide the full-field in 0.2s, while the high-fidelity models require 15m. Ac concerns the fluid dynamics test cases, the ROM achieves a maximum speed up $1 \times {10}^5$ compared to the numerical solver, providing the unsteady velocity components in 0.25s.

Future steps amongst others include the implementation of the proposed algorithm to estimate the uncertainty quantification of the material properties and boundary conditions in structural mechanics and thermal simulations in the field of aerospace and marine. We further motivate readers to incorporate physics during the training of the DL models, aiming to improve the generalizability of the non-intrusive ROMs.   

\section{Acknowledgements}
We gratefully acknowledge Mr. C. Kokkinos, Dr. D. Pettas and Dr. D. Lampropoulos (FEAC Engineering P.C.) for their insightful remarks on the implementation of the numerical test cases, and Prof. F. Kopsaftopoulos (Department of Mechanical, Aerospace, and Nuclear Engineering, Rensselaer Polytechnic Institute, Troy) for the useful discussions regarding the evolution of the time series. This work has been further supported by the development of teaching for the installation and operation of wind turbines-computer-aided modeling  (DAAD) project.

\section*{Appendix}
\addcontentsline{toc}{section}{Appendices}
\renewcommand{\thesubsection}{\Alph{subsection}}
\subsection{Neural architecture search} \label{NAS}
The implementation of the neural architecture search, used for the optimum investigation of the activation functions for the FFNN, are presented in this chapter. The results are displayed for each benchmark problem separately.
\begin{table}[H]
\centering{
\begin{tabular}{|c|c|c|c|c|c|}
\hline
\diagbox[width=\dimexpr \textwidth/8+2\tabcolsep\relax, height=1.1cm]{\boldsymbol{$\varepsilon_{mse}$}}{$\textbf{function}$}  & $$\textbf{Sigmoid}$$ & $$\textbf{Leaky ReLU}$$ & $$\textbf{ReLU}$$ & $$\textbf{ELU}$$ & $$\textbf{Swish}$$ \\ \hline
$$\textbf{Training loss}$$       & $1.7\times10^{-4}$       & $5.8\times10^{-5}$          & $6.6\times10^{-5}$    & $3.7\times10^{-5}$   & $2.7\times10^{-4}$     \\ \hline
$$\textbf{Validation loss}$$     & $1.3\times10^{-4}$       & $7.3\times 10^{-5}$           & $2.1\times10^{-4}$   & $4.9\times10^{-4}$   & $1.9\times10^{-4}$     \\ \hline
\end{tabular} \caption{Training and validation $\bm{\varepsilon_{mse}}$ loss of the FFNN model through each activation function for the CD equation test case.}}
\end{table}
\setlength{\parskip}{0pt}

\begin{table}[H]
\centering{
\begin{tabular}{|c|c|c|c|c|c|}
\hline
\diagbox[width=\dimexpr \textwidth/8+2\tabcolsep\relax, height=1.1cm]{\boldsymbol{$\varepsilon_{mse}$}}{$\textbf{function}$}  & $$\textbf{Sigmoid}$$ & $$\textbf{Leaky ReLU}$$ & $$\textbf{ReLU}$$ & $$\textbf{ELU}$$ & $$\textbf{Swish}$$ \\ \hline
$$\textbf{Training loss}$$       & $3.1\times10^{-5}$       & $1.2\times10^{-5}$          & $1.6\times10^{-3}$    & $1.2\times10^{-4}$   & $1.9\times10^{-5}$     \\ \hline
$$\textbf{Validation loss}$$     & $6.4\times10^{-4}$       & $1.1\times10^{-4}$           & $8.7\times10^{-5}$   & $1.1\times10^{-3}$   & $1.0\times10^{-3}$     \\ \hline
\end{tabular} \caption{Training and validation $\bm{\varepsilon_{mse}}$ error of the FFNN model through each activation function for the flow around a cylinder test case.}}
\end{table}
\hspace{-0.1cm}
\begin{table}[H]
\centering{
\begin{tabular}{|c|c|c|c|c|c|}
\hline
\diagbox[width=\dimexpr \textwidth/8+2\tabcolsep\relax, height=1.1cm]{\boldsymbol{$\varepsilon_{mse}$}}{$\textbf{function}$}  & $$\textbf{Sigmoid}$$ & $$\textbf{Leaky ReLU}$$ & $$\textbf{ReLU}$$ & $$\textbf{ELU}$$ & $$\textbf{Swish}$$ \\ \hline
$$\textbf{Training loss}$$       & $6.1 \times 10^{-6}$       & $5.1\times 10^{-6}$          & $3.9\times10^{-6}$    & $4.1\times10^{-6}$   & $8.7\times 10^{-6}$     \\ \hline
$$\textbf{Validation loss}$$     & $6.4\times10^{-6}$       & $1.4\times10^{-5}$           & $5.4\times10^{-6}$   & $5.8\times10^{-6}$   & $5.7\times10^{-6}$     \\ \hline
\end{tabular} \caption{Training and validation $\bm{\varepsilon_{mse}}$ error of the FFNN model through each activation function for the blood flow in the arterial segment test case.}}
\end{table}
\setlength{\parskip}{10pt}

\subsection{Neural network architectures} \label{NN arch}
The architecture of the neural networks, utilized in the current work for each benchmark problem, are presented in this chapter.
\begin{table}[H]
\centering{
\begin{tabular}{|cc|cc|}
\hline
\multicolumn{2}{|c|}{\textbf{Encoder}}             & \multicolumn{2}{c|}{\textbf{Decoder}}           \\ \hline
\multicolumn{2}{|c|}{\textbf{Activation function}} & \multicolumn{2}{c|}{ELU}                        \\ \hline
\multicolumn{2}{|c|}{\textbf{Kernel size}}         & \multicolumn{2}{c|}{(3 x 3)}                    \\ \hline
\multicolumn{1}{|c|}{\textbf{Layer}} & \textbf{Output shape} & \multicolumn{1}{c|}{\textbf{Layer}} & \textbf{Output shape} \\ \hline
\multicolumn{1}{|c|}{Input}        & (16,16,1)     & \multicolumn{1}{c|}{Input}       & 4            \\ \hline
\multicolumn{1}{|c|}{Conv2D}       & (16, 16, 25)  & \multicolumn{1}{c|}{Dense}       & 10           \\ \hline
\multicolumn{1}{|c|}{Max pooling}  & (8, 8, 25)    & \multicolumn{1}{c|}{Dense}       & 20           \\ \hline
\multicolumn{1}{|c|}{Conv2D}       & (8, 8, 10)    & \multicolumn{1}{c|}{Dense}       & 12           \\ \hline
\multicolumn{1}{|c|}{Max pooling}    & (4, 4, 10)            & \multicolumn{1}{c|}{Reshape}        & (2, 2, 3)             \\ \hline
\multicolumn{1}{|c|}{Flatten}      & 160           & \multicolumn{1}{c|}{Conv2D}      & (2, 2, 10)   \\ \hline
\multicolumn{1}{|c|}{Dense}        & 20            & \multicolumn{1}{c|}{Up sanpling} & (4, 4, 10)   \\ \hline
\multicolumn{1}{|c|}{Dense}        & 10            & \multicolumn{1}{c|}{Conv2D}      & (4, 4, 25)   \\ \hline
\multicolumn{1}{|c|}{Dense}        & 4             & \multicolumn{1}{c|}{Up sampling} & (8, 8, 25)   \\ \hline
\multicolumn{1}{|c|}{-}            & -             & \multicolumn{1}{c|}{Conv2D}      & (8, 8, 30)   \\ \hline
\multicolumn{1}{|c|}{-}            & -             & \multicolumn{1}{c|}{Up sampling} & (16, 16, 30) \\ \hline
\multicolumn{1}{|c|}{-}            & -             & \multicolumn{1}{c|}{Conv2D}      & (16, 16, 1)  \\ \hline
\end{tabular}\caption{CAE architecture for the CD equation test case.}}
\end{table}
\setlength{\parskip}{0pt}
\begin{table}[H]
\centering{
\begin{tabular}{|c|c|}
\hline
\textbf{Layer} & \textbf{Output shape} \\ \hline
Input          & (10, 6)               \\ \hline
LSTM           & (10, 50)              \\ \hline
LSTM           & (10, 50)              \\ \hline
LSTM           & 50                   \\ \hline
Dense          & 4                     \\ \hline
\end{tabular}\caption{LSTM architecture for the CD test equation case.}}
\end{table}
\begin{table}[H]
\centering{
\begin{tabular}{|c|c|}
\hline
\textbf{Activation function} & Leaky ReLU            \\ \hline
\textbf{Layer}               & \textbf{Output shape} \\ \hline
Input                        & 3                     \\ \hline
Dense                        & 50                    \\ \hline
Dense                        & 4                     \\ \hline
\end{tabular}\caption{FFNN architecture for the CD equation test case.}}
\end{table}

\begin{table}[H]
\centering{
\begin{tabular}{|cc|cc|}
\hline
\multicolumn{2}{|c|}{\textbf{Encoder}}             & \multicolumn{2}{c|}{\textbf{Decoder}}           \\ \hline
\multicolumn{2}{|c|}{\textbf{Activation function}} & \multicolumn{2}{c|}{ELU}                        \\ \hline
\multicolumn{2}{|c|}{\textbf{Kernel size}}         & \multicolumn{2}{c|}{(3 x 3)}                    \\ \hline
\multicolumn{1}{|c|}{\textbf{Layer}} & \textbf{Output shape} & \multicolumn{1}{c|}{\textbf{Layer}} & \textbf{Output shape} \\ \hline
\multicolumn{1}{|c|}{Input}        & (32, 32, 1)     & \multicolumn{1}{c|}{Input}       & 4            \\ \hline
\multicolumn{1}{|c|}{Conv2D}       & (32, 32, 30)  & \multicolumn{1}{c|}{Dense}       & 10           \\ \hline
\multicolumn{1}{|c|}{Max pooling}  & (16, 16, 30)    & \multicolumn{1}{c|}{Dense}       & 25           \\ \hline
\multicolumn{1}{|c|}{Conv2D}       & (16, 16, 20)    & \multicolumn{1}{c|}{Dense}       & 50           \\ \hline
\multicolumn{1}{|c|}{Max pooling}    & (8, 8, 20)            & \multicolumn{1}{c|}{Dense}        & 12            \\ \hline
\multicolumn{1}{|c|}{Conv2D}      & (8, 8, 15)           & \multicolumn{1}{c|}{Reshape}      & (2, 2, 3)   \\ \hline
\multicolumn{1}{|c|}{Max pooling}        & (4, 4, 15)            & \multicolumn{1}{c|}{Conv2D} & (2, 2, 10 )   \\ \hline
\multicolumn{1}{|c|}{Conv2D}        & (4, 4, 10)            & \multicolumn{1}{c|}{Up sampling}      & (4, 4, 10)   \\ \hline
\multicolumn{1}{|c|}{Max pooling}        & (2, 2, 10)             & \multicolumn{1}{c|}{Conv2D} & (4, 4, 15)   \\ \hline
\multicolumn{1}{|c|}{Dense}            & 40             & \multicolumn{1}{c|}{Up sampling}      & (8, 8, 15)   \\ \hline
\multicolumn{1}{|c|}{Dense}            & 50             & \multicolumn{1}{c|}{Conv2D} & (8, 8, 20) \\ \hline
\multicolumn{1}{|c|}{Dense}            & 25             & \multicolumn{1}{c|}{Up sampling}      & (16, 16, 20)  \\ \hline
\multicolumn{1}{|c|}{Dense}            & 10             & \multicolumn{1}{c|}{Conv2D}      & (16, 16, 30)  \\ \hline
\multicolumn{1}{|c|}{Dense}            & 4             & \multicolumn{1}{c|}{Up sampling}      & (32, 32, 30)  \\ \hline
\multicolumn{1}{|c|}{-}            & -             & \multicolumn{1}{c|}{Conv2D}      & (32, 32, 1)  \\ \hline
\end{tabular}\caption{CAE architecture for the flow around a cylinder test case.}}
\end{table}
\begin{table}[H]
\centering{
\begin{tabular}{|c|c|}
\hline
\textbf{Activation function} & Sigmoid           \\ \hline
\textbf{Layer}               & \textbf{Output shape} \\ \hline
Input                        & 2                     \\ \hline
Dense                        & 32                    \\ \hline
Dense                        & 32                    \\ \hline
Dense                        & 32                    \\ \hline
Dense                        & 4                     \\ \hline
\end{tabular}\caption{FFNN architecture for the flow around a cylinder test case.}}
\end{table}

\begin{table}[H]
\centering{
\begin{tabular}{|c|c|}
\hline
\textbf{Layer} & \textbf{Output shape} \\ \hline
Input          & (10, 5)               \\ \hline
LSTM           & (10, 30)              \\ \hline
LSTM           & (10, 30)              \\ \hline
LSTM           & (10, 30)              \\ \hline
LSTM           & 30                   \\ \hline
Dense          & 4                     \\ \hline
\end{tabular}\caption{LSTM architecture for the flow around a cylinder test case.}}
\end{table}

\begin{table}[H]
\centering{
\begin{tabular}{|cc|cc|}
\hline
\multicolumn{2}{|c|}{\textbf{Encoder}}             & \multicolumn{2}{c|}{\textbf{Decoder}}           \\ \hline
\multicolumn{2}{|c|}{\textbf{Activation function}} & \multicolumn{2}{c|}{ELU}                        \\ \hline
\multicolumn{2}{|c|}{\textbf{Kernel size}}         & \multicolumn{2}{c|}{(3 x 3)}                    \\ \hline
\multicolumn{1}{|c|}{\textbf{Layer}} & \textbf{Output shape} & \multicolumn{1}{c|}{\textbf{Layer}} & \textbf{Output shape} \\ \hline
\multicolumn{1}{|c|}{Input}        & (16, 16, 3)     & \multicolumn{1}{c|}{Input}       & 4            \\ \hline
\multicolumn{1}{|c|}{Conv2D}       & (16, 16, 30)  & \multicolumn{1}{c|}{Dense}       & 10           \\ \hline
\multicolumn{1}{|c|}{Max pooling}  & (8, 8, 30)    & \multicolumn{1}{c|}{Dense}       & 40           \\ \hline
\multicolumn{1}{|c|}{Conv2D}       & (8, 8, 20)    & \multicolumn{1}{c|}{Dense}       & 12           \\ \hline
\multicolumn{1}{|c|}{Max pooling}    & (4, 4, 20)            & \multicolumn{1}{c|}{Reshape}        & (2, 2, 3)             \\ \hline
\multicolumn{1}{|c|}{Conv2D}    & (4, 4, 10)            & \multicolumn{1}{c|}{Conv2D}        & (2, 2, 10)             \\ \hline
\multicolumn{1}{|c|}{Max pooling}    & (2, 2, 10)            & \multicolumn{1}{c|}{Up sampling}        & (4, 4, 10)             \\ \hline
\multicolumn{1}{|c|}{Flatten}      & 40           & \multicolumn{1}{c|}{Conv2D}      & (4, 4, 20)   \\ \hline
\multicolumn{1}{|c|}{Dense}        & 40            & \multicolumn{1}{c|}{Up sanpling} & (8, 8, 20)   \\ \hline
\multicolumn{1}{|c|}{Dense}        & 10            & \multicolumn{1}{c|}{Conv2D}      & (8, 8, 30)   \\ \hline
\multicolumn{1}{|c|}{Dense}        & 4             & \multicolumn{1}{c|}{Up sampling} & (16, 16, 30)   \\ \hline
\multicolumn{1}{|c|}{-}            & -             & \multicolumn{1}{c|}{Conv2D}      & (16, 16, 3)   \\ \hline 
\end{tabular}\caption{CAE architecture for the blood flow in the arterial segment.}}
\end{table}
\begin{table}[H]
\centering{
\begin{tabular}{|c|c|}
\hline
\textbf{Activation function} & ReLU            \\ \hline
\textbf{Layer}               & \textbf{Output shape} \\ \hline
Input                        & 2                     \\ \hline
Dense                        & 32                     \\ \hline
Dense                        & 32                    \\ \hline
Dense                        & 32                    \\ \hline
Dense                        & 4                     \\ \hline
\end{tabular}\caption{FFNN architecture for the blood flow in the arterial segment.}}
\end{table}

\begin{table}[H]
\centering{
\begin{tabular}{|c|c|}
\hline
\textbf{Layer} & \textbf{Output shape} \\ \hline
Input          & (30, 5)               \\ \hline
LSTM           & (30, 80)              \\ \hline
LSTM           & (30, 80)              \\ \hline
LSTM           & (30, 80)              \\ \hline
LSTM           & 80                    \\ \hline
Dense          & 4                     \\ \hline
\end{tabular}} \caption{LSTM architecture for the blood flow in the arterial segment.}
\end{table}

\bibliography{references}  
\end{document}